\documentclass{fairmeta}

\usepackage{times}

\usepackage{amsmath,amsfonts,bm}

\def\eqref#1{equation~\ref{#1}}

\def\1{\bm{1}}

\DeclareMathAlphabet{\mathsfit}{\encodingdefault}{\sfdefault}{m}{sl}
\SetMathAlphabet{\mathsfit}{bold}{\encodingdefault}{\sfdefault}{bx}{n}

\usepackage{booktabs}
\usepackage{colortbl}
\usepackage{flafter}
\usepackage{caption}
\usepackage{titletoc}
\usepackage{enumitem}
\usepackage{subcaption}
\usepackage{float}
\usepackage{graphicx}
\usepackage{hyperref}
\usepackage{longtable}
\usepackage{microtype}
\usepackage{multirow}
\usepackage{placeins}


\usepackage{tabularx}
\usepackage{wrapfig}
\usepackage{xcolor}
\usepackage{url}
\usepackage{tikz}

\usetikzlibrary{arrows.meta,calc,fit,positioning,shapes.geometric}

\definecolor{moea3b}{HTML}{C44E52}   
\definecolor{moea10b}{HTML}{6A4C9C}  
\definecolor{moelavender}{HTML}{9C9FC4}
\definecolor{moeseafoam}{HTML}{78A99A}
\definecolor{moemauve}{HTML}{A88FAF}
\definecolor{moesteel}{HTML}{7897C2}
\definecolor{moeneutral}{HTML}{969696}
\definecolor{moelavenderdark}{HTML}{6F739A}
\definecolor{moeseafoamdark}{HTML}{4F806F}
\definecolor{moemauvedark}{HTML}{806783}
\definecolor{moetablegray}{HTML}{F1F2F4}
\definecolor{moetablehighlight}{HTML}{F3F2FC}

\definecolor{linkblue}{HTML}{2E74B5}
\hypersetup{
  colorlinks = true,
  citecolor  = linkblue,   
  linkcolor  = linkblue,   
  urlcolor   = linkblue,   
  filecolor  = linkblue,
}
\definecolor{plmteal}{HTML}{D1E2DD}
\definecolor{plmindigo}{HTML}{D0D9EB}
\definecolor{plmpurple}{HTML}{DAD1ED}
\definecolor{plmmagenta}{HTML}{E5D1EB}
\definecolor{plmtealdark}{HTML}{005953}
\definecolor{plmindigodark}{HTML}{324779}
\definecolor{plmpurpledark}{HTML}{5A3477}
\definecolor{plmgray}{HTML}{F3F3F3}
\usepackage{pifont}

\newcommand{\method}{\textsc{MoEMB}}

\title{\method{}: Scaling Universal Multimodal Embeddings with Efficient Mixture-of-Experts Models}

\author[2, *]{Xuanming Cui}
\author[1]{Shlok Kumar Mishra}
\author[1]{Wentao Bao}
\author[1]{Aashu Singh}
\author[1]{Zihao Wang}
\author[1]{Xiangjun Fan}
\author[1]{Jun Xiao}
\author[2, \dagger]{Ser-Nam Lim}
\author[1, \dagger]{Jianpeng Cheng}

\affiliation[1]{AI at Meta}
\affiliation[2]{University of Central Florida}

\contribution[*]{Work done at Meta}
\contribution[\dagger]{Equal advising}

\abstract{
Universal multimodal embedding (UME) increasingly demands encoder's capacity for handling a broad range
of tasks and modalities with increased complexity. Prior scaling methods either increase the representation size, retrieval effort, or scales the encoder into a heavy multimodal LLM.  Recent works, such as Think-Then-Embed (TTE), explore scaling via reasoning tokens. However, embedding models are hard to scale up: increasing parameters directly trades off against the large training batch size that contrastive learning needs, and retrieval has to be served under tight latency. Moreover, UME tasks are diverse in complexity, where scaling up embedders can bring significantly more redundant computation for simple retrievals. In this work, we propose \method{}, which instead scales UME along the
\emph{expert} axis through mixture-of-experts (MoE), growing encoder capacity
while preserving non-autoregressive encoding. Through a
systematic study of the design space and training recipes for MoE-based UME,
\method{} sets a new state of the art on both MMEB-V2 and MRMR among models trained on
public MMEB-family data: with only $\sim$3B active parameters, \method{} surpasses TTE-based methods with $>4\times$ active parameters, using significantly less compute. To further improve the scalability and efficiency, we conduct the first
comprehensive study of adaptive computation for MoE-based embedding, spanning
diverse strategies across training-based and inference-only methods. Together,
these results support expert scaling as an effective and efficient direction for
UME, with adaptive computation further improving efficiency for MLLM-based embedding models towards real-world, large-scale retrieval systems.
}
\date{September 7, 2026}

\begin{document}

\maketitle

\section{Introduction}
\label{sec:introduction}

Embedding models map heterogeneous inputs into a shared space for 
retrieval. While earlier embedding tasks focus primarily on simple symmetric retrieval, recent universal multimodal embedding (UME) tasks now require a single encoder to handle diverse modalities, with diverse tasks spanning classification, question answering, retrieval, grounding,
and document search~\citep{meng2025vlm2vecv2}.

\begin{wrapfigure}{r}{0.51\textwidth}
  \vspace{-35pt}
  \centering
  \includegraphics[width=0.98\linewidth]{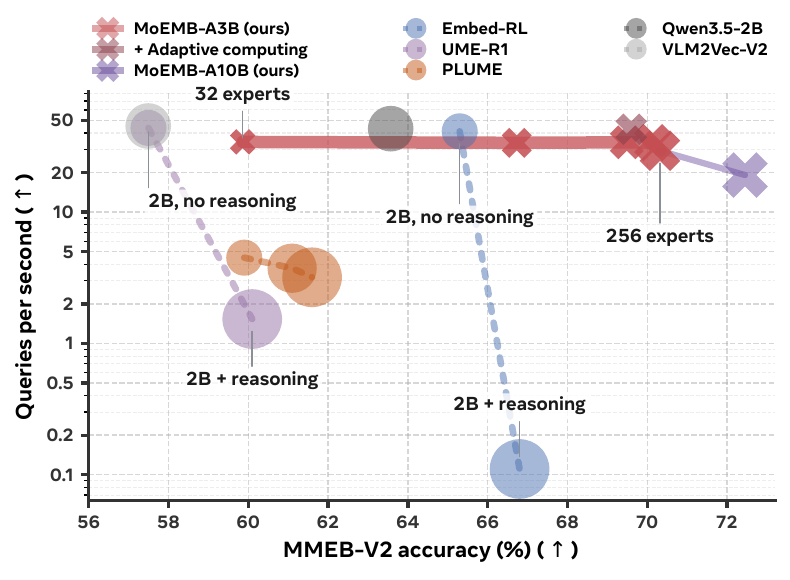}
  \vspace{-11pt}
  \captionsetup{font=footnotesize}
  \caption{ \textbf{Accuracy vs.\ QPS.}  Scaling via MoE
({\color{moea3b}\method{}}) raises accuracy by $>10\%$ at near-constant
throughput. Details can be found in ~\ref{sec:fig1-details}.}
\label{fig:flops-acc}
\vspace{-23pt}
\end{wrapfigure}

\begin{figure}[!t]
  \centering
  \includegraphics[width=\textwidth]{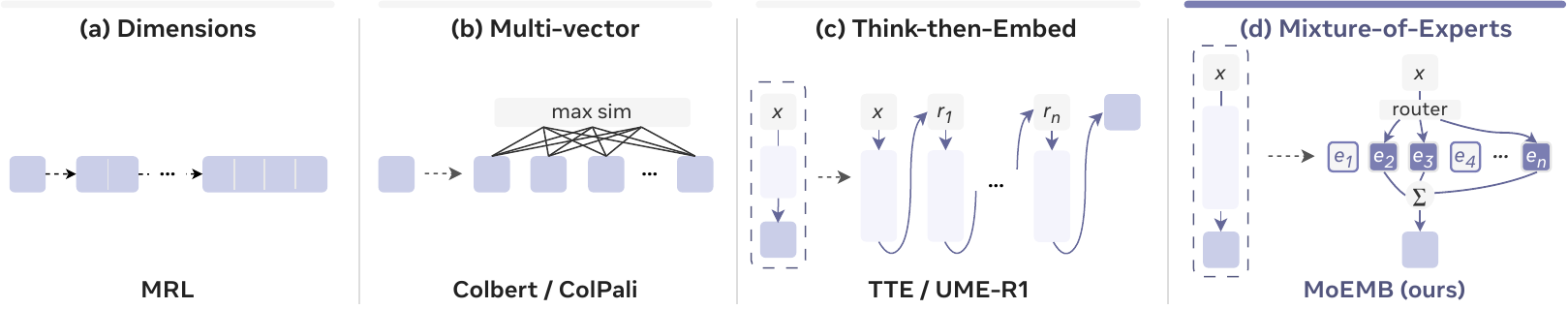}
  \vspace{-16pt}
  \caption{\textbf{Four ways to scale multimodal embeddings.}  Dimension (a) and
  multi-vector (b) methods scale the output representation; TTE-based
  methods (c) scale via intermediate reasoning tokens.  \method{} (d) instead scales the
  encoder along the expert axis through sparse MoE.}
  \label{fig:scaling-taxonomy}
  \vspace{-15pt}
\end{figure}

As the embedding tasks grow
more varied and complex, performance depends more on the encoder's capacity for visual reasoning and
instruction following. However, most work on scaling embeddings still targets the stored representation instead of that capacity (Fig.~\ref{fig:scaling-taxonomy}).  Matryoshka
Representation Learning adjusts the embedding
dimensionality~\citep{kusupati2022matryoshka}, while multi-vector retrieval methods~\citep{colbert, faysse2025colpali, xiao2025metaembed} keep
multiple vectors per input for finegrained retrieval.  These methods scale storage and
retrieval budget, but they leave the encoder itself
untouched, and so cannot supply the understanding and reasoning that harder UME
tasks demand.

\begin{wrapfigure}{r}{0.34\textwidth}

  \centering
  \vspace{-10pt}
  \includegraphics[width=\linewidth]{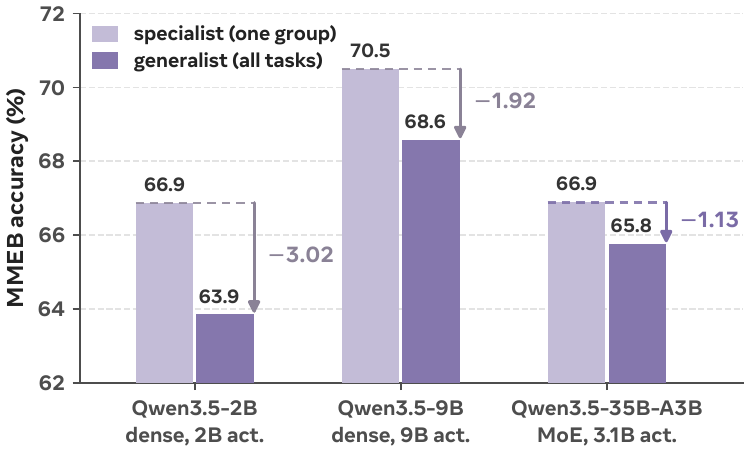}
  \vspace{-16pt}
  \captionsetup{font=footnotesize}
  \caption{Performance degradation for switching from specialist to generalist shrinks with model’s total capacity. Each model trained on the full MMEB
mixture (generalist) vs. one task alone (specialist).
Details in App.~\ref{sec:capacity-ablation}.}
  \label{fig:capacity-ablation}
  \vspace{-1.0\baselineskip}
\end{wrapfigure}

To scale the encoder itself, a line of work turns multimodal large language
models (MLLMs) into embedders, and these now lead UME
benchmarks~\citep{jiang2025vlm2vec,lin2025mmembed,li2026qwen3vlembedding}. However, even at larger scales than CLIP-based~\citep{radford2021clip} lightweight models, these MLLM-based embedders still struggle to cover the diverse
tasks of UME. We empirically find (Fig.~\ref{fig:capacity-ablation}) that one
embedder trained jointly over all tasks is outperformed by the same model
trained per task; Qwen3-VL-Embedding likewise reports degraded
MTEB~\citep{mteb} performance against text-only
Qwen3-Embedding~\citep{qwen3embedding}. Two explanations are possible:
destructive gradient interference between tasks, or a model simply not capacious
enough to serve them all at once.  Both require further scaling up the
embedder's capacity.

A recent line of think-then-embed (TTE) methods scale for reasoning-heavy
retrieval by generating intermediate reasoning tokens before producing the final
embedding~\citep{cui2025ifmtte,lan2025umer1,cheng2026tteflash}. However, these methods improve performance with a significant cost of computation. For instance, on the text
partition of MMEB-V3~\citep{mmebv3}, it takes 1.7 days for a regular MLLM-based 2B encoder~\citep{meng2025vlm2vecv2} to complete evaluation, while UME-R1-2B~\citep{lan2025umer1}, a TTE-based model, needs roughly \textbf{7.8 H200-days} to encode queries, and \textbf{431 H200-days} for candidates, which is prohibitively expensive for real-world deployment.

In general, scaling a UME encoder is difficult: (1) search and recommendation systems are required to
embed astronomically many queries and items, under low latency; (2) contrastive
training encodes and backpropagates through both the query and the target, so the activation memory is doubled; (3) added parameters trade against
batch size, a key component of contrastive quality; and (4) UME tasks vary
widely in complexity: for simpler tasks the extra LLM backbone
may be wasteful, since the MLLM's own contrastively pretrained vision
encoder~\citep{radford2021clip} could already suffice.

To address challenges (1--3), we propose \method{}, the first
Mixture-of-Experts (MoE) UME model. Instead of scaling the embedding dimension, dense parameters, or reasoning tokens, we scale the encoder along the
\emph{expert} axis. Sparse MoE is well studied for language
modeling~\citep{jiang2024mixtralexperts, kimiteam2026kimik3openfrontier} but
remains underexplored for embedding models, for which we fill the gap. We study three approaches
to obtain an MoE-based embedder: (1) adapting a pretrained native sparse MoE, (2)
dense-to-MoE upcycling, and (3) modality-level MoE; to our knowledge, \method{}
is the first to adapt and scale pretrained sparse multimodal MoE backbones for
UME. Expert scaling gains over 10 points at negligible additional compute
(Fig.~\ref{fig:flops-acc}): with only $\sim$3.1B active parameters,
\method{}-A3B reaches a new SOTA of 70.4 on MMEB-V2, surpassing all UME models trained on the
MMEB V2 dataset, including TTE-based encoders that require significantly more
computes.


Next, to tackle issue (4), heterogeneous complexity across UME tasks, we conduct
the first comprehensive study of adaptive computation for MoE-based UME
embedders. Prior adaptive computation primarily focus on text
generation~\citep{huang2025dynamicllava, chen2024fastv, bai2025diep} and
traditional vision tasks like classification~\citep{yin2022avit}, with little work on embeddings. Moreover, these works typically study
one mechanism at a time. In this work, we study five actions --- token pruning, layer skipping, expert
skipping, adaptive top-$k$, and expert pruning --- under both training-based and inference-only
settings. Token pruning and adaptive top-$k$ give the strongest accuracy--compute trade-off, and
inference-only variants are competitive with their trained counterparts.
Applying them saves notable compute at under one point of accuracy.

\noindent\textbf{Our contributions are twofold.}

\noindent\textbf{1. Expert scaling for UME.} \textbf{(a)} We introduce expert capacity as a scaling axis for UME and conduct the first
comprehensive study of its model designs and training recipes. \textbf{(b)} Without any special data mining or sophisticated multi-stage training, \method{} sets a new SOTA on two large-scale benchmarks, MMEB-V2 and
MRMR~\citep{zhang2025mrmr}, among models trained only on public MMEB-family data, surpassing TTE-based
encoders that consume significantly more compute.

\noindent\textbf{2. Adaptive computation for MoE-based UME.} \textbf{(a)} We conduct the first systematic exploration for adaptive computation for MoE-based UME embedders, across five actions and both training-based and inference-only methods. \textbf{(b)} By leveraging adaptive computation methods, we substantially reduce forward compute for $< 1\%$ performance degradation. \textbf{(c)} We are the first to explore token pruning based on Gated Delta Networks' $\beta$ as a strong alternative to attention scores. \textbf{(d)} We provide corresponding inference implementation for realizing saved computes into wall-clock saving.



\section{Preliminaries}
\label{sec:preliminaries}

A UME embedder maps a multimodal, instruction-conditioned query or target
$\{q, t\}=(I_x, I_v, I_t)$ composed of a task
instruction $I_x$, optionally visual inputs $I_v$ and a text input $I_t$, to a single
vector~\citep{meng2025vlm2vecv2}.  We build it from a pretrained multimodal
LLM: one encoder $F_\theta$ processes both queries $t$ and targets $t$:
\begin{equation*}
e_\theta(\{q, t\})=\operatorname{norm}\big(\operatorname{Pool}(F_\theta(\{q, t\}))\big),
\end{equation*}
where $\operatorname{norm}$ is the normalization operation, and $\operatorname{Pool}$ is the pooling operation that produces the final embedding representation~\citep{cui2025ifmtte}. For a batch of $N$ pairs, we follow the standard one-directional InfoNCE
objective~\citep{jiang2025vlm2vec}:
\begin{equation}
\mathcal L_{\mathrm{emb}}=-\frac{1}{N}\sum_i
\log\frac{e^{S_{ii}}}{\sum_j e^{S_{ij}}}, \quad S_{ij}=\operatorname{cos}(q_i,t_j)/\tau
\label{eq:infonce}
\end{equation}
where $\tau$ denotes the temperature, and $\operatorname{cos}$ is the cosine similarity function.

\section{Building Sparse MoE UME Models}
\label{sec:expert-scaling}

\subsection{MoE Backbone Designs}

We explore three ways
to build a sparse MoE-based embedder: \textbf{(1) Native MoE:} we reuse a backbone whose experts and router are learned in standard language model
pretraining; \textbf{(2) Dense-to-sparse Upcycling:} we convert a
dense MLLM into sparse MoE during contrastive training;
\textbf{(3) modality-level routing:} we perform routing on the \emph{modality} level.


\textbf{(1) Native MoE.}
\label{sec:native-moe}
We directly reuse an MLLM whose feed-forward
blocks are already sparse MoE, composed of a shared expert $\mathrm{FFN}_{\mathrm{shared}}$ and $E$ sparse experts. For a token's hidden state $h\in\mathbb{R}^{d}$,
the pretrained router $W_r\in\mathbb{R}^{E\times d}$ scores all $E$ experts and keeps the $k$
highest, $\mathcal S=\operatorname{TopK}(W_r h,\,k)$.  Their scores are renormalized over
$\mathcal S$, while a learned vector $w_{\mathrm{shared}}\in\mathbb{R}^{d}$ scales the shared expert:
\begin{equation}
\operatorname{MoE}(h)=\gamma\,\mathrm{FFN}_{\mathrm{shared}}(h)
+\sum_{j\in\mathcal S}\alpha_j\,\mathrm{FFN}_j(h),
\qquad
\begin{aligned}
\alpha_j &= \tfrac{e^{(W_r h)_j}}{\sum_{u\in\mathcal S}e^{(W_r h)_u}},\\[3pt]
\gamma &= \operatorname{sigmoid}\big(w_{\mathrm{shared}}\,h\big).
\end{aligned}
\label{eq:moe}
\end{equation}
This variant retains MoE pretrained from language modeling. Empirically we find this setting leads to the best performance. We therefore use it as the default setting for all subsequent experiments.


\textbf{(2) Dense-to-MoE Upcycling.}
\label{sec:upcycle}
We explore upcycling a dense MLLM into a sparse MoE embedder by splitting
its feed-forward block and replicating the pieces into $E$ experts, then learn a
router over them from scratch. We empirically find its performance to lag behind the native MoE setting. Therefore, we report its best variant in Tab.~\ref{tab:moe-designs} and provide further details in App.~\ref{sec:upcycling}.

\begin{wrapfigure}{r}{0.41\textwidth}
  \centering
  \includegraphics[width=\linewidth]{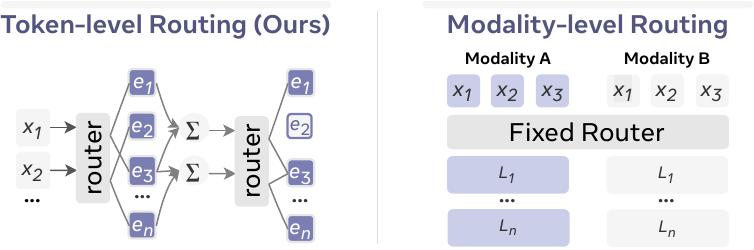}
  \vspace{-13pt}
  \captionsetup{font=footnotesize}
  \caption{Token- vs.\ modality-level routing.  
  }
  \label{fig:routing-illustration}
  \vspace{-15pt}
\end{wrapfigure}
\textbf{(3) Modality-level Experts.}
\label{sec:task-modality}
As shown in Fig.~\ref{fig:routing-illustration}, we explore an alternative routing setting, where we replicate the dense feed-forward blocks into modality-level experts, where a fixed router forwards all tokens from one modality to its corresponding expert. This is in contrast to the regular \emph{token-level routing}, where each token is dynamically routed to its top-$k$ experts. We find this variant slightly improves over the dense baseline, but lags behind the native MoE setting (Tab.~\ref{tab:moe-designs}).

\subsection{Pooling, router adaptation, and training dynamics}
\label{sec:pooling-router}

Beyond determining an effective, MoE backbone design, and adapting a pretrained MoE model to a contrastive embedding objective also requires other design and training choices. In this work we explore three aspects: \textit{pooling}, \textit{router adaptation}, and \textit{training dynamics} for MoE-UME.

\textbf{Pooling Strategy.} Previous UME works~\citep{cui2025ifmtte} have discovered that pooling via last token often outperforms other pooling methods like mean~\citep{geminiembedding} or latent attention pooling~\citep{nvembed}. Building on this observation, we only compare two designs: a simple last token pooling, versus a designated \texttt{<emb>} token. Empirically we find the latter to perform better (Tab.~\ref{tab:recipe}).

\textbf{Router Adaptation and Auxiliary Objectives.}
Because pre-trained routers establish delicate token-to-expert assignments, downstream contrastive fine-tuning can destabilize learned expert allocation. To systematically evaluate router stability versus adaptability, we compare frozen and trainable router configurations alongside three auxiliary objectives: (1) \emph{load-balancing and $z$-losses} to prevent expert collapse; (2) a \emph{conflict-aware loss} to mitigate cross-task gradient conflict by penalizing routing distribution overlap across conflicting tasks; and (3) \emph{task conditioning}, which incorporates discrete task labels or continuous instruction embeddings alongside token hidden states into router logit computation. Full details for each formulation are provided in App.~\ref{sec:routing-objectives}. Empirically, we find that the simple frozen router instead yields the best performance, which indicates that the
routers learned during language pre-training are also near-optimal for the contrastive objective (Tab.~\ref{tab:recipe}).

\textbf{Adaptation Dynamics.} We systematically ablate the hyperparameter settings governing contrastive adaptation for MoE-based UME: global batch size, contrastive temperature, LoRA rank, learning rate, and training epochs (Fig.~\ref{fig:hyperparam}).


\begin{figure}[t]
  \centering
  \includegraphics[width=\textwidth]{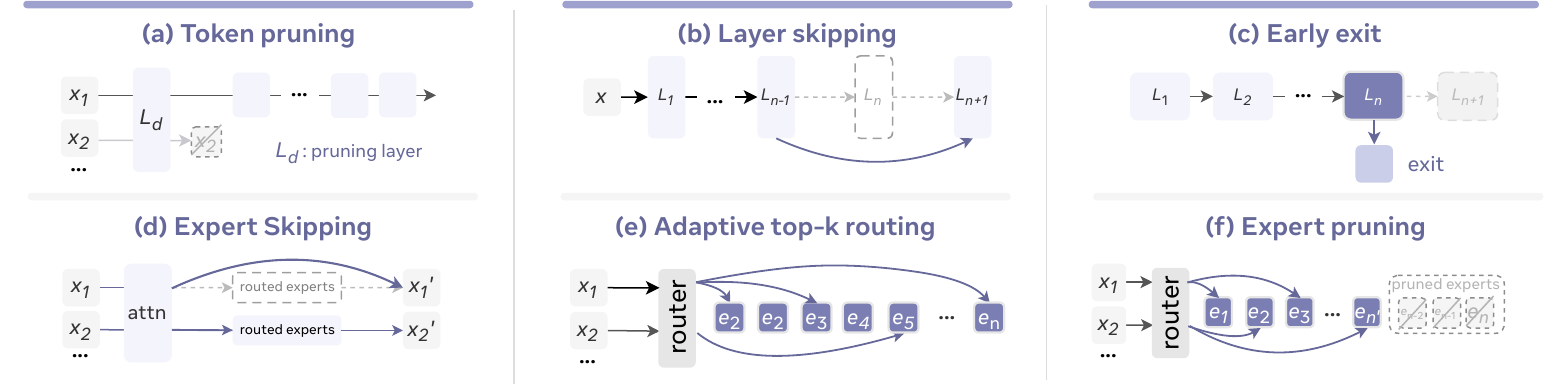}
  \caption{We consider six adaptive computing actions in \method{}. The top-three (a-c) are general adaptive computing actions, while the bottom-three (d-f) are specific to MoEs.}
  \label{fig:efficiency-actions}
\end{figure}

\section{Adaptive Computation for MoE-UME}
\label{sec:adaptive-compute}

\begin{table}[t]
\centering
\caption{ Summarization for different designs.  \emph{Train} is training-based, \emph{Inf.}\ means inference-only.  \emph{Basis} names the work a method
is adapted from, or \emph{ours} where we know of no prior counterpart.
$^{\dagger}$marks those detailed in
Apps.~\ref{sec:additional-efficiency} and~\ref{sec:failed-adaptivity}
instead of in the main text.}
\label{tab:efficiency-design}
\setlength{\tabcolsep}{4pt}
\renewcommand{\arraystretch}{1.12}
\resizebox{\textwidth}{!}{%
\begin{tabular}{l l c l}
\toprule
\textbf{Action} & \textbf{Method} & \textbf{Regime} & \textbf{Basis} \\
\midrule
\multirow{3}{*}{Token pruning}
 & One-shot pruner after decoder block $n$ (Gumbel/STE)      & Train & \citet{huang2025dynamicllava} \\
 & Layer-wise gradual pruning with GRPO$^{\dagger}$ & Train & ours \\
 & GDN $\beta$-ranked one-shot drop                          & Inf.  & ours \\
\midrule
\multirow{3}{*}{Expert skipping}
 & Per-token three-way skip gate (soft fade)                 & Train & ours \\
 & GDN $\beta$-ranked expert skip                            & Inf.  & ours \\
 & Per-layer Otsu split on $\beta$ or router logit$^{\dagger}$ & Inf.  & \citet{otsu1979threshold} \\
\midrule
\multirow{3}{*}{Adaptive top-$k$}
 & Null experts, load-balance--scheduled$^{\dagger}$         & Train & \citet{zeng2024adamoe} \\
 & Dynamic modality thresholding                             & Inf.  & \citet{huang2025modes} \\
 & Fixed router-score threshold$^{\dagger}$                  & Inf.  & ours \\
\midrule
Layer skip
 & Trained layer-skip gate (STE, distillation)$^{\dagger}$   & Train & ours \\
\midrule
Early exit
 & Multi-depth contrastive loss               & Train & ours \\
\midrule
\multirow{2}{*}{Expert pruning}
 & Routing frequency, router probability mass                & Inf.  & \citet{lu2024notallexperts} \\
 & MAN, MSAN, REAP saliency                                  & Inf.  & \citet{liu2026scoreexperts,lasby2025reap} \\
\bottomrule
\end{tabular}}
\vspace{-10pt}
\end{table}

While sparse MoE improves the quality--compute tradeoff, it still runs a fixed per-token compute budget with potential computation redundancy. To further improve efficiency and scalability for MoE-based UME models, we explore various adaptive computation actions. Fig.~\ref{fig:efficiency-actions} illustrates the six types of adaptive computing actions we study, covering both training-based (\S\ref{sec:training-efficiency}) and inference-only (\S\ref{sec:inference-eff}) settings. Tab.~\ref{tab:efficiency-design} summarizes different methods we studied. Due to the space limit, we discuss the most effective methods in the following sections, and leave the rest to App.~\ref{sec:additional-efficiency} and~\ref{sec:failed-adaptivity}.

\subsection{Training-Based Adaptive Computation Methods}
\label{sec:training-efficiency}

\textbf{Token pruning.} We study two types of token pruning: \emph{one-shot} and \emph{gradual} pruning.
A \emph{one-shot} pruning makes a single pruning decision at an early
layer~\citep{huang2025dynamicllava} $K$; a
\emph{gradual} pruning applies a per-token gate at every layer, so a token's alive
mass $\alpha_{t,\ell}=\prod_{j<\ell}(1-p^{\mathrm{drop}}_{t,j})$ decays with depth
and its downstream compute is removed progressively.  We train one-shot pruning with a straight-through estimator (STE). For gradual pruning, we explore both STE and an AViT-style~\citep{yin2022avit} halting mechanism, where both methods failed due to unstable training (App.~\ref{sec:failed-adaptivity}). Therefore, we realize gradual pruning with the GRPO policy of App.~\ref{sec:controller-details}, which samples the discrete action instead of differentiating through it.  Both settings distill from a separate teacher embedding obtained without adaptive computing: $e^{\mathrm{full}}$. Given the target budget $\rho^{\star}$ and the model's current keep rate  $\rho_{\mathrm{keep}}$, the overall loss becomes:
\begin{equation*}
\mathcal{L}_{\mathrm{drop}} \;=\; \mathcal{L}_{\mathrm{emb}} \;+\;
\lambda_{\mathrm{d}}\big(1-\cos(e,\,e^{\mathrm{full}})\big) \;+\;
\lambda_{\mathrm{b}}\big(\rho_{\mathrm{keep}} - \rho^{\star}\big)^{2}.
\end{equation*} 

\textbf{Learned expert skipping.}
Inspired by Mixture-of-Depths~\citep{raposo2024mixtureofdepths}, we add a skip gate $p^{\mathrm{skip}}_{t,\ell}\in[0,1]$ over existing routers, which determines whether to skip the entire experts computation:
\begin{equation*}
\mathcal{L}_{\mathrm{skip}} \;=\; \mathcal{L}_{\mathrm{emb}} \;+\;
\lambda_{\mathrm{c}}\big(1 - \rho_{\mathrm{skip}}\big),
\qquad
\rho_{\mathrm{skip}} \;=\; \frac{1}{L\,|\mathcal{T}|}\sum_{\ell=1}^{L}\sum_{t\in\mathcal{T}}
p^{\mathrm{skip}}_{t,\ell},
\end{equation*}
with $L$ being the number of MoE layers, $|\mathcal{T}|$ denoting the number of tokens, and $p^{\mathrm{skip}}_{t,\ell}$ the normalized probability produced by each skip gate. The skip budget $\rho_{\mathrm{skip}}$ applies to the entire model, which allows the skip rate to differ across layers (Fig.~\ref{fig:trained-skip}). As a hard skip gate is non-differentiable, we train it through a soft fade. Let $u_{t,\ell}$ be the output of the routed experts for token $t$ at layer $\ell$, and $v_{t,\ell}$ the rest of that layer's output; the routed output $h_{t,\ell}$ is scaled by the non-skip probability $1-p^{\mathrm{skip}}_{t,\ell}$:
\begin{equation*}
h_{t,\ell} \;=\; v_{t,\ell} \;+\; \big(1-p^{\mathrm{skip}}_{t,\ell}\big)\,u_{t,\ell}.
\end{equation*}


\textbf{Early exit.} We supervise a set of candidate exit depths $\mathcal{D}$ within the LLM backbone.  Let $q_d,t_d$ denote the query and target embeddings taken at depth $d$, and $\bar q_L,\bar t_L$ the full-depth embeddings, precomputed from same pre-trained embedder. The objective has two parts, $\mathcal{L}_{\mathrm{exit}}=\mathcal{L}_{\mathrm{con}}+\lambda_{\mathrm{a}}\mathcal{L}_{\mathrm{dis}}$: the mean contrastive loss at every depth, and a distillation loss pulling every embedding into a shared space onto its cached teacher:
\begin{equation*}
\mathcal{L}_{\mathrm{con}} \;=\; \frac{1}{|\mathcal{D}^{+}|}\sum_{d\in\mathcal{D}^{+}}
\mathcal{L}_{\mathrm{emb}}(q_d,t_d),
\qquad
\mathcal{L}_{\mathrm{dis}} \;=\; \frac{1}{2|\mathcal{D}|}\sum_{d\in\mathcal{D}}
\Big[\big(1-\cos(q_d,\bar q_L)\big)+\big(1-\cos(t_d,\bar t_L)\big)\Big],
\end{equation*}
where $\mathcal{L}_{\mathrm{emb}}(\cdot,\cdot)$ is Eq.~\ref{eq:infonce} evaluated on the given
pair of embedding sets.


\subsection{Inference-Based Efficiency}
\label{sec:inference-eff}

\textbf{Token importance without an attention map.}
Inference-only token pruning conventionally ranks visual tokens by the attention mass
they receive~\citep{chen2024fastv}, which presumes the attention matrix is
available.  However, as the efficient FlashAttention kernels~\citep{flashattention} do not materialize the
attention map, we need to find an alternative training-free signal for token importance.

In this work, we propose to leverage the writing strength $\beta$ in  Gated DeltaNet (GDN)~\citep{yang2024gateddelta}, which is used in recent frontier MLLMs~\citep{qwenteam2026qwen35,kimiteam2026kimik3openfrontier}. At each GDN
layer the gated delta rule updates a recurrent memory $S$ using a 
token-dependent write strength.  Within one head of one GDN layer, the
rule keeps a memory $S_t\in\mathbb{R}^{d_v\times d_k}$ and updates it as
\begin{equation}
S_{t} \;=\; S_{t-1}\big(\alpha_{t}(I - \beta_{t}\,k_{t}k_{t}^{\top})\big)
\;+\; \beta_{t}\,v_{t}k_{t}^{\top},
\qquad
\beta_{t} \;=\; \operatorname{sigmoid}\big(\mathbf{W}_{\beta}^{\top}x_{t}\big)\in(0,1),
\label{eq:gdn-beta}
\end{equation}
where $x_t$ is the token's hidden state, $k_t\in\mathbb{R}^{d_k}$ and
$v_t\in\mathbb{R}^{d_v}$ its $\ell_2$-normalized key and its value, $\alpha_t\in(0,1)$ the forget gate, and $I$ the
identity.
Thus,
$\beta_{t}$ interpolates between leaving the memory untouched ($\beta_t{\to}0$)
and fully overwriting its $k_{t}$ direction with $v_{t}$ ($\beta_t{\to}1$), which provides a natural surrogate for token importance. Averaging $\beta_t$ over those layers and over heads gives each token an importance score, which drives every inference-only token and expert decision below.

\textbf{Token pruning.}
We utilize the $\beta$ score to prune visual tokens. After decoder block $K$ we rank
the visual tokens by $\beta_t$, keep the highest-scoring fraction, and
discard the rest for every layer that follows.

\textbf{Adaptive top-$k$.}
Let $\alpha_{\ell t(1)} \ge \dots \ge \alpha_{\ell t(k)}$ denote the sorted routing weights for token $t$ at layer $\ell$ (Eq.~\ref{eq:moe}). Adaptive top-$k$ routing selects a dynamic number of experts $k_{\ell t} \le k$ per token via a predefined condition. For instance, MoDES~\citep{huang2025modes} keeps an expert $m$ if its scaled routing weight satisfies $s_{\ell} \alpha_{\ell t(m)} \ge \tau$, for which the global threshold $\tau$ and layer-scaling factors $s_{\ell}$ are estimated during a calibration pass (App.~\ref{sec:modes-calibration}). Here, $\tau$ acts as a uniform decision threshold, while $s_{\ell}$ upweights critical layers to retain more experts where capacity matters the most.


\textbf{Total expert pruning.} We consider pruning experts from the model permanently, which
reduces memory instead of compute.  Each layer keeps the $E'$ experts with the largest saliency
$\phi_j$, measured on a calibration pass. We compare two families of saliency.
\emph{Router-based} scores read the router alone~\citep{lu2024notallexperts}, either as the
fraction of calibration tokens routed to expert $j$ (routing frequency) or as its mean router
probability. \emph{Output-based} scores instead read the output produced by the expert, averaged over the tokens
routed to it,
\begin{equation}
\phi_j \;=\; \operatorname*{mean}_{t\,\rightarrow\,j}\, w_{tj},
\qquad
w_{tj}\in\big\{\underbrace{\lVert o_{tj}\rVert}_{\text{MAN}},\;
\underbrace{\lVert o_{tj}\rVert^{2}}_{\text{MSAN}},\;
\underbrace{\alpha_{\ell tj}\lVert o_{tj}\rVert}_{\text{REAP}}\big\},
\qquad
o_{tj}=\mathrm{FFN}_j(h_{\ell t}),
\label{eq:prune-scores}
\end{equation}
where $o_{tj}$ is the expert's output and $\alpha_{\ell tj}$ its routing weight from
Eq.~\ref{eq:moe}.  The three output-based scores differ only in how they weight that output.
MAN and MSAN~\citep{liu2026scoreexperts} take its norm and squared norm, ignoring the router
entirely, so an expert is judged purely by how large an output it produces.
REAP~\citep{lasby2025reap} additionally scales the norm by the routing weight. Empirically, we find this variant to generally perform better (Fig.~\ref{fig:budget-accuracy}).

\section{Experiments}
\label{sec:experiments}

\subsection{Experimental Setup}
\label{sec:setup}

\textbf{Models and training.}
We adapt the Qwen3.5-35B-A3B and Qwen3.5-122B-A10B multimodal
MoEs~\citep{qwenteam2026qwen35}.  All training is done with two epochs at length 8192 in
BF16, global batch 512, AdamW at learning rate $10^{-4}$ with ten warmup steps and
cosine decay, and LoRA of rank 64 and scale 128. We use Nvidia Megatron Core MoE~\citep{megatron} with Expert Parallelism for accelerated MoE-UME training. App.~\ref{sec:implementation-details} gives full implementation details.

\begin{table}[t]\centering\scriptsize\setlength{\tabcolsep}{2.4pt}\renewcommand{\arraystretch}{0.95}
\captionsetup{font=footnotesize}
\caption{ MMEB-V2 results. \textbf{bold}/\underline{underline} mark best/second within the public-data group.  $^{\dagger}$Embed-RL pairs a shared 8.77B autoregressive reasoner with a 4.44B/2.13B embedder, giving 13.2B/10.9B active parameters.  $^{\ddagger}$These models do not disclose an active parameter count.  CLS: classification, QA: question answering, RET: retrieval, GD: grounding, MRT: moment retrieval; V1/V2: ViDoRe-v1/v2, VR: VisRAG, OOD: out-of-domain.}\label{tab:main-results}
\vspace{-5pt}
\resizebox{\textwidth}{!}{\begin{tabular}{ll *{4}{c}c *{4}{c}c *{4}{c}c c}
\toprule
\multirow{2}{*}{\textbf{Model}} & \multirow{2}{*}{\textbf{\shortstack[l]{Act.\\(B)}}} & \multicolumn{5}{c}{\textbf{Image}} & \multicolumn{5}{c}{\textbf{Video}} & \multicolumn{5}{c}{\textbf{VisDoc}} & \multirow{2}{*}{\textbf{All}}\\
\cmidrule(lr){3-7}\cmidrule(lr){8-12}\cmidrule(lr){13-17}
\hspace{1.1em} & & CLS & QA & RET & GD & \textbf{Ov} & CLS & QA & RET & MRT & \textbf{Ov} & V1 & V2 & VR & OOD & \textbf{Ov} & \\
\midrule
\textbf{\# Datasets}$\rightarrow$ & & 10 & 10 & 12 & 4 & 36 & 5 & 5 & 5 & 3 & 18 & 10 & 4 & 6 & 4 & 24 & 78\\
\midrule
\rowcolor{plmgray}\multicolumn{18}{c}{\textbf{Models adapted with external data}}\\
\midrule
\multicolumn{18}{@{}l}{\hspace{0.4em}\textbf{\textit{$\sim$2--4B active params}}}\\
\hspace{1.1em}DME-2B & 2.2 & 73.0 & 73.8 & 76.2 & 90.7 & 76.3 & 82.2 & 57.6 & 53.5 & 45.6 & 61.3 & 86.0 & 59.3 & 92.5 & 72.0 & 80.9 & 74.2\\
\hspace{1.1em}Qwen3-VL-Embedding-2B & 2.1 & 70.3 & 74.3 & 74.8 & 88.5 & 75.0 & 71.9 & 64.9 & 53.9 & 53.3 & 61.9 & 84.4 & 65.3 & 86.4 & 69.4 & 79.2 & 73.2\\
\hspace{1.1em}RzenEmbed-v1-2B & 2.2 & 65.3 & 61.7 & 73.8 & 77.8 & 68.5 & 45.6 & 47.5 & 38.3 & 36.7 & 42.6 & 87.0 & 57.6 & 85.4 & 67.1 & 78.4 & 65.6\\
\hspace{1.1em}Ops-MM-Embedding-v1-2B & 2.2 & 68.1 & 65.1 & 69.2 & 80.8 & 69.0 & 53.6 & 55.6 & 41.7 & 33.7 & 47.6 & 76.4 & 53.2 & 77.6 & 65.1 & 71.0 & 64.7\\
\multicolumn{18}{@{}l}{\hspace{0.4em}\textbf{\textit{$\ge$8B active params}}}\\
\hspace{1.1em}DME-Large$^{\ddagger}$ & -- & 75.4 & 82.9 & 79.6 & 95.2 & 81.1 & 90.2 & 75.7 & 64.5 & 62.2 & 74.4 & 89.1 & 61.5 & 93.8 & 75.1 & 83.4 & 80.2\\
\hspace{1.1em}Octen-VL-Embedding-Large$^{\ddagger}$ & -- & 75.7 & 84.5 & 80.6 & 94.4 & 81.9 & 87.1 & 82.1 & 68.0 & 60.4 & 76.0 & 85.9 & 66.3 & 86.6 & 72.2 & 80.5 & 80.1\\
\hspace{1.1em}DME-Medium & 9.4 & 74.5 & 80.9 & 78.2 & 94.5 & 79.8 & 87.7 & 71.0 & 61.0 & 58.5 & 70.8 & 87.6 & 57.8 & 94.5 & 73.5 & 82.0 & 78.4\\
\hspace{1.1em}Qwen3-VL-Embedding-8B & 8.1 & 74.2 & 81.1 & 80.2 & 92.3 & 80.1 & 78.4 & 71.0 & 58.7 & 56.1 & 67.1 & 87.2 & 69.9 & 88.7 & 73.3 & 82.4 & 77.8\\
\hspace{1.1em}IFM-TTE-7B & 8.3 & 76.7 & 78.5 & 74.6 & 89.3 & 77.9 & 60.5 & 67.9 & 51.7 & 54.9 & 59.2 & 85.2 & 71.5 & 92.8 & 67.3 & 81.8 & 74.8\\
\hspace{1.1em}WeMM-Embedding-8B & 8.8 & 73.5 & 76.1 & 78.6 & 92.9 & 78.1 & 66.5 & 71.7 & 56.4 & 55.2 & 63.2 & 89.5 & 59.3 & 90.4 & 35.1 & 75.6 & 73.9\\
\hspace{1.1em}RzenEmbed-v2-7B & 8.3 & 70.6 & 71.7 & 78.5 & 92.1 & 75.9 & 58.8 & 63.5 & 51.0 & 45.5 & 55.7 & 89.7 & 60.7 & 88.7 & 69.4 & 81.2 & 72.9\\
\hspace{1.1em}Ops-MM-Embedding-v1-7B & 8.3 & 69.6 & 69.6 & 73.1 & 87.2 & 72.7 & 59.7 & 62.2 & 45.7 & 43.2 & 53.8 & 80.0 & 59.6 & 79.3 & 67.5 & 74.4 & 68.9\\
\hspace{1.1em}E5-Omni-7B & 8.0 & 66.7 & 68.5 & 73.0 & 83.9 & 71.2 & 46.6 & 52.9 & 36.7 & 34.2 & 43.5 & 87.6 & 62.4 & 87.5 & 34.6 & 74.5 & 65.8\\
\hspace{1.1em}GME-Qwen2-VL-7B & 8.3 & 57.6 & 34.6 & 71.2 & 59.5 & 55.9 & 37.3 & 50.3 & 28.3 & 37.5 & 38.4 & 89.6 & 55.5 & 85.0 & 69.4 & 79.4 & 59.1\\
\midrule
\rowcolor{plmgray}\multicolumn{18}{c}{\textbf{Models adapted only with public MMEB-family data}}\\
\midrule
\multicolumn{18}{@{}l}{\hspace{0.4em}\textbf{\textit{$\sim$2--4B active params}}}\\
\hspace{1.1em}PLUME & 2.1 & \underline{66.5} & 59.2 & 67.6 & 79.7 & 66.3 & 45.0 & 52.3 & 33.5 & 46.7 & 44.1 & 72.1 & 49.8 & 78.1 & 57.4 & 67.5 & 61.6\\
\hspace{1.1em}UME-R1-2B & 2.2 & 64.8 & 62.8 & 67.6 & 77.2 & 66.6 & 44.3 & 50.9 & 32.9 & 39.7 & 42.2 & 72.4 & 46.2 & 79.2 & 57.8 & 67.3 & 61.2\\
\hspace{1.1em}VLM2Vec V2-2B & 2.2 & 62.9 & 56.4 & 69.6 & 77.1 & 64.9 & 39.2 & 34.7 & 28.4 & 37.5 & 34.7 & 74.4 & 44.6 & 79.3 & 62.5 & 68.7 & 59.1\\
\hspace{1.1em}BToks & 2.2 & 64.3 & 59.8 & 68.8 & 77.4 & 66.0 & 43.7 & 47.0 & 33.0 & 33.6 & 39.9 & 71.1 & 38.6 & 81.3 & 38.1 & 62.7 & 59.0\\
\hspace{1.1em}VLM2Vec V1-2B & 2.2 & 58.6 & 49.2 & 65.0 & 73.0 & 59.7 & 33.3 & 30.7 & 20.4 & 30.7 & 28.5 & 20.6 & 13.2 & 52.2 & 46.9 & 31.7 & 43.9\\
\rowcolor{plmteal}\hspace{1.1em}\textbf{\method{}-A3B} & 3.1 & 60.6 & \underline{73.8} & 72.4 & \underline{92.4} & 71.8 & 53.8 & \underline{65.9} & \underline{45.3} & 45.8 & \underline{53.5} & \underline{85.0} & \underline{57.8} & \underline{88.5} & \underline{82.8} & \underline{81.0} & \underline{70.4}\\
\multicolumn{18}{@{}l}{\hspace{0.4em}\textbf{\textit{$\ge$8B active params}}}\\
\hspace{1.1em}Embed-RL-4B$^{\dagger}$ & 13.2 & 63.7 & 70.5 & 71.3 & 91.3 & 71.2 & \underline{57.6} & 58.4 & 45.1 & \textbf{49.5} & 53.0 & 80.2 & 53.4 & 84.9 & 67.1 & 74.7 & 68.1\\
\hspace{1.1em}Embed-RL-2B$^{\dagger}$ & 10.9 & 62.8 & 67.9 & 68.6 & 90.4 & 69.2 & 57.0 & 55.9 & 45.1 & 49.4 & 52.1 & 79.9 & 52.0 & 84.6 & 65.7 & 74.1 & 66.8\\
\hspace{1.1em}UME-R1-7B & 8.3 & \textbf{67.1} & 69.2 & 71.9 & 84.9 & 71.2 & 48.6 & 60.7 & 38.2 & 39.3 & 47.5 & 75.7 & 50.5 & 83.7 & 58.3 & 70.6 & 65.6\\
\hspace{1.1em}UniME-V2 & 8.0 & 65.6 & 68.7 & \underline{73.1} & 90.8 & \underline{71.8} & 37.2 & 50.6 & 28.9 & 39.6 & 39.0 & 61.8 & 42.0 & 70.5 & 58.3 & 60.1 & 60.6\\
\hspace{1.1em}VLM2Vec V1-7B & 8.3 & 62.8 & 56.5 & 69.4 & 81.9 & 65.4 & 39.0 & 30.1 & 29.1 & 39.2 & 33.8 & 20.0 & 9.2 & 58.9 & 57.1 & 34.1 & 48.5\\
\rowcolor{plmteal}\hspace{1.1em}\textbf{\method{}-A10B} & 10.2 & 64.4 & \textbf{75.4} & \textbf{74.7} & \textbf{92.7} & \textbf{74.0} & \textbf{58.1} & \textbf{67.6} & \textbf{47.3} & \underline{49.4} & \textbf{56.3} & \textbf{85.5} & \textbf{62.7} & \textbf{89.1} & \textbf{82.9} & \textbf{82.2} & \textbf{72.4}\\
\bottomrule\end{tabular}}
\end{table}

\textbf{Benchmarks.} We evaluate our models on two comprehensive and large-scale multimodal retrieval benchmarks: MMEB V2~\citep{meng2025vlm2vecv2} and MRMR~\citep{zhang2025mrmr}. \textbf{MMEB-V2}~\citep{meng2025vlm2vecv2} comprises 78 tasks spanning image, video, and visual document (VisDoc) retrieval tasks. \textbf{MRMR}~\citep{zhang2025mrmr} is a reasoning-intensive multimodal retrieval benchmark, which requires multi-step logical inference and complex domain-specific interpretation across multimodal contexts (e.g., cross-modal knowledge retrieval, theorem identification, and contradiction detection). 

\subsection{Results and Ablations on MoE Design and Training Dynamics for MoE-UME}
\label{sec:results-moe}
\label{sec:scaling}
\vspace{-5pt}

\textbf{Main results.} Tab.~\ref{tab:main-results} shows that \method{} achieves superior performance on MMEB V2, outperforming all public-data baselines across model scales despite using fewer active parameters and avoiding costly autoregressive reasoning passes. As shown in Fig.~\ref{fig:flops-acc}, expert scaling yields a significantly better accuracy--throughput trade-off than TTE-based methods: expanding total expert capacity delivers substantial accuracy gains with negligible throughput loss. Furthermore, on MRMR (Tab.~\ref{tab:mrmr}), \method{}-A3B achieves the top overall score despite its smaller active footprint, outperforming TTE methods designed for reasoning-intensive retrieval. This confirms that scaling model capacity via MoE is more effective than test-time reasoning scaling. Moreover, as MMEB V2 is largely OOD, many frontier models rely on external corpora mirroring evaluation formats, inflating scores through data leakage. On evaluation splits where MMEB V2 provides in-domain supervision (e.g., ViDoRe and VisRAG), \method{} achieves SOTA performance without external adaptation data, matching or exceeding significantly larger models trained on massive external data and complex pipelines.

\begin{wraptable}{r}{0.52\textwidth}
\vspace{-10pt}
\centering
\captionsetup{font=footnotesize}
\caption{\footnotesize Ablation on different expert scaling design.}\label{tab:moe-designs}
\vspace{-8pt}
\footnotesize\setlength{\tabcolsep}{4pt}\renewcommand{\arraystretch}{1.15}
\resizebox{\linewidth}{!}{%
\begin{tabular}{lrcccc}
\toprule
\textbf{Method} & \textbf{Act.} & \textbf{Image} & \textbf{Video} & \textbf{VisDoc} & \textbf{Overall} \\
\midrule
Qwen3.5-2B (dense)          & 2.2B & 64.4 & 43.0 & 77.7 & 63.6 \\
Dense-to-MoE upcycling &   3.0B & 62.6 & 45.3 & 77.3 & 63.1 \\
Modality-wise MoE  & 2.2B & 65.1 & 43.6 & 78.3 & 64.2 \\
\rowcolor{plmteal}\textbf{\method{}-A3B (native)} & 3.1B & \textbf{71.8} & \textbf{53.5} & \textbf{81.0} & \textbf{70.4} \\
\bottomrule
\end{tabular}}
\vspace{-5pt}
\end{wraptable}
\textbf{Ablations on MoE Design.}
We compare different MoE designs at equal active compute (Tab.~\ref{tab:moe-designs}). We can observe upcycling (App.~\ref{sec:upcycling}) does not outperform the dense backbone. We hypothesize this is due to that the limited scale for contrastive training over the MMEB V2 dataset is not able to train a strong router. Meanwhile, modality-wise MoE shows only marginal gains over the dense baseline. In contrast, fine-tuning a native pretrained sparse MoE backbone delivers substantial gains.

\begin{table}[htbp]\centering
\caption{Reasoning-intensive retrieval on MRMR~\citep{zhang2025mrmr}. $^{\dagger}$reported by the MRMR benchmark; other rows are our evaluations.  \textbf{Bold}/\underline{underline} mark best/second per column.}\label{tab:mrmr}
\vspace{-5pt}
\setlength{\tabcolsep}{4pt}\renewcommand{\arraystretch}{1.12}
\resizebox{0.9\textwidth}{!}{%
\begin{tabular}{lc cccc cccc ccc c}
\toprule
& & \multicolumn{4}{c}{\textbf{Knowledge}} & \multicolumn{4}{c}{\textbf{Theorem}} & \multicolumn{3}{c}{\textbf{Contradiction}} & \\
\cmidrule(lr){3-6}\cmidrule(lr){7-10}\cmidrule(lr){11-13}
\textbf{Model} & \textbf{Act. (B)} & Art & Med. & Sci. & Hum. & Math & Phy. & Eng. & Bus. & Neg. & Design & Traffic & \textbf{Avg} \\
\midrule
\multicolumn{14}{@{}l}{\textit{$\le$4B active parameters}}\\
\midrule
BToks & 2.2 & 72.0 & 48.2 & 62.2 & 62.3 & 22.5 & 36.1 & 29.1 & 49.0 & 14.5 & 46.4 & 36.0 & 43.5 \\
UME-R1-2B & 2.2 & 70.8 & 48.9 & 60.6 & 57.8 & 21.3 & 31.5 & 27.2 & 47.0 & 11.5 & 33.7 & 30.1 & 40.0 \\
PLUME & 2.1 & 69.9 & 47.7 & 63.6 & 56.7 & 18.6 & 30.3 & 27.7 & 44.5 & 9.5 & 43.2 & 27.5 & 39.9 \\
VLM2Vec-V2 & 2.2 & 53.1 & 28.9 & 43.4 & 39.9 & 11.7 & 22.3 & 22.5 & 43.0 & 11.0 & 8.3 & 25.0 & 28.1 \\
ColPali$^{\dagger}$ & 3.0 & 36.1 & 29.9 & 42.7 & 29.2 & 7.3 & 17.5 & 13.5 & 34.6 & \textbf{28.5} & 19.4 & 18.2 & 25.2 \\
VISTA$^{\dagger}$ & 0.2 & 21.3 & 27.8 & 32.6 & 17.0 & 18.8 & 17.1 & 17.3 & 28.6 & \underline{20.0} & 20.2 & 9.4 & 20.9 \\
VLM2Vec$^{\dagger}$ & 4.0 & 53.5 & 22.4 & 36.7 & 24.0 & 2.1 & 2.8 & 2.8 & 2.9 & 11.5 & 5.6 & 18.3 & 18.1 \\
\rowcolor{plmteal}\textbf{\method{}-A3B} & 3.1 & \textbf{80.3} & \textbf{62.6} & \textbf{73.3} & \textbf{76.3} & 28.2 & 35.3 & 34.3 & \textbf{54.0} & 11.0 & 63.6 & \textbf{49.2} & \textbf{51.6} \\
\midrule
\multicolumn{14}{@{}l}{\textit{$\ge$7B active parameters}}\\
\midrule
LaME-7B & 8.3 & 72.7 & 57.1 & 69.8 & 63.3 & 29.9 & \underline{44.4} & \textbf{38.0} & 49.4 & 6.5 & \textbf{66.9} & 39.8 & 48.9 \\
UME-R1-7B & 8.3 & 79.2 & 58.2 & 71.8 & 66.1 & 28.2 & 40.2 & 32.7 & 50.3 & 6.0 & 57.6 & 39.5 & 48.2 \\
Ops-MM-Embedding$^{\dagger}$ & 7.0 & \underline{79.3} & 52.5 & 70.0 & 67.8 & 27.7 & 39.5 & 30.1 & 52.3 & 8.0 & 55.9 & \underline{45.8} & 48.1 \\
Embed-RL-4B & 13.2 & 73.5 & 53.1 & 60.5 & 69.8 & \textbf{32.9} & \textbf{45.6} & 35.1 & \underline{52.6} & 6.0 & 54.3 & 31.7 & 46.8 \\
UniME-V2-7B & 8.0 & 71.0 & 49.7 & 58.4 & 59.6 & 25.1 & 39.8 & 28.4 & 51.0 & 7.0 & 37.0 & 34.6 & 42.0 \\
MM-Embed$^{\dagger}$ & 8.0 & 65.6 & 53.0 & 63.5 & 62.8 & 23.6 & 30.8 & 27.4 & 44.9 & 7.0 & 23.8 & 34.9 & 39.8 \\
GME-Qwen2-VL$^{\dagger}$ & 7.0 & 54.3 & 40.1 & 46.8 & 45.6 & 28.8 & 36.0 & 30.2 & 45.1 & 15.0 & 26.3 & 29.6 & 36.2 \\
E5-V$^{\dagger}$ & 8.0 & 25.1 & 11.7 & 16.6 & 10.8 & 2.1 & 3.4 & 2.5 & 5.2 & 11.5 & 3.7 & 2.1 & 8.6 \\
\bottomrule\end{tabular}}
\end{table}

\begin{figure}[htbp]\centering
    \vspace{-5pt}
  \includegraphics[width=\textwidth]{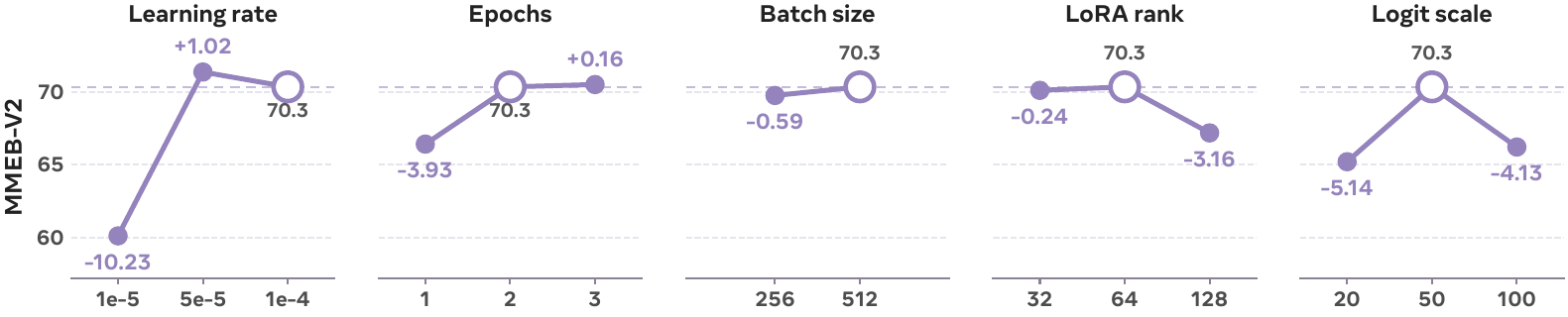}
  \vspace{-15pt}
  \caption{Hyperparameter ablations on the 35B backbone.  Each panel
  varies one setting.}
  \vspace{-10pt}
  \label{fig:hyperparam}
\end{figure}



\begin{wraptable}{r}{0.47\textwidth}
\centering
\captionsetup{font=footnotesize}
\caption{Ablations on pooling designs and router adaptations for MoE-UME.  Details for each router setting are provided in App.~\ref{sec:routing-objectives}.}
\vspace{-8pt}
\label{tab:recipe}
\footnotesize
\setlength{\tabcolsep}{4pt}
\renewcommand{\arraystretch}{1.1}
\resizebox{\linewidth}{!}{%
\begin{tabular}{llccccc}
\toprule
\multirow{2}{*}{\textbf{Pooling}} & \multirow{2}{*}{\textbf{Routing objective}}
 & \textbf{Router} & \multirow{2}{*}{\textbf{Image}} & \multirow{2}{*}{\textbf{Video}}
 & \multirow{2}{*}{\textbf{VisDoc}} & \multirow{2}{*}{\textbf{Overall}} \\
 & & \textbf{trained} & & & & \\
\midrule
\emph{last}  & \ding{55}         & \ding{51} & 70.7 & 51.6 & 81.1 & 69.50 \\
\emph{last}  & LB\,$+$\,$z$      & \ding{51} & 65.6 & 53.7 & 77.8 & 66.62 \\
\emph{last}  & conflict-aware    & \ding{51} & 70.4 & 53.2 & 81.3 & 69.80 \\
\emph{\textless emb\textgreater} & \ding{55} & \ding{51} & 71.1 & 54.5 & 81.1 & 70.34 \\
\emph{\textless emb\textgreater} & task-ID prior     & \ding{51} & 69.9 & 53.5 & 81.0 & 69.55 \\
\emph{\textless emb\textgreater} & instruction prior & \ding{51} & 70.1 & 54.1 & 80.3 & 69.55 \\
\emph{\textless emb\textgreater} & \ding{55} & \ding{55} & \textbf{71.8} & 53.5 & 81.0 & \textbf{70.38} \\
\bottomrule
\end{tabular}}
\vspace{-10pt}
\end{wraptable}
\textbf{Ablations on Pooling and Routing Designs.} Evaluating adaptation choices on the native sparse backbone (Tab.~\ref{tab:recipe}) reveals three insights:
\textbf{(1) Pooling Strategy}: Replacing last-token pooling with a dedicated \texttt{<emb>} token provides notable performance gain.
\textbf{(2) Router Updating}: Freezing the pretrained router performs identically to fine-tuning.
\textbf{(3) Routing Objectives}: auxiliary load-balancing and $z$-losses degrade accuracy, while task priors and conflict-aware routing offer no meaningful gains.
Together, (2) and (3) indicate that routing assignments learned during language pre-training remain highly effective for contrastive adaptation.

\textbf{Ablations on Training Hyperparameters.} Fig.~\ref{fig:hyperparam} shows ablations on major training hyperparameters, where we can draw three observations: \textbf{(1)} the effective range is narrow for learning rate, where small decreases improve accuracy, but larger drops degrade it sharply due to underfitting; \textbf{(2)} performance peaks at two epochs, after which contrastive gains saturate. \textbf{(3)} logit scale is the most sensitive parameter, while batch size has the least impact.


\subsection{Results and Analysis for Adaptive Computation on MoE-UME}
\label{sec:results-adaptive}
\label{sec:inference-results}

\begin{table}[htbp]\centering
\vspace{-12pt}
\caption{\footnotesize Summarized results on adaptive-computation. We show a representative budget for each method.}
\vspace{-5pt}
\label{tab:inference}
\setlength{\tabcolsep}{5pt}\renewcommand{\arraystretch}{1.2}
\resizebox{\textwidth}{!}{%
\begin{tabular}{lll ccccc c}
\toprule
\textbf{Action} & \textbf{Method} & \textbf{Budget} & \textbf{Image} & \textbf{Video} & \textbf{VisDoc} & \textbf{Overall} & \textbf{FLOPs$\downarrow$\%} & \textbf{QPS$\uparrow\!\times$} \\
\midrule
--- & No reduction & top-8 / keep all & 71.0 & 54.5 & 81.1 & 70.3 & --- & 1.00 \\
\midrule
\multicolumn{9}{@{}l}{\textit{Training-based methods}}\\
\midrule
Adaptive top-$k$ & AdaMoE\,{\scriptsize(App.~\ref{sec:trained-arms})} & 2.94 experts & 68.2\,{\scriptsize($-$2.8)} & 49.1\,{\scriptsize($-$5.4)} & 79.3\,{\scriptsize($-$1.8)} & 67.2\,{\scriptsize($-$3.1)} & 15 & 1.07 \\
Expert pruning & DiEP~\citep{bai2025diep} & $E{=}128$ & 66.1\,{\scriptsize($-$4.9)} & 47.9\,{\scriptsize($-$6.6)} & 79.9\,{\scriptsize($-$1.2)} & 66.2\,{\scriptsize($-$4.1)} & 0.3 & 1.00 \\
Expert skip & Learned skip gate & skip 50\% & 69.7\,{\scriptsize($-$1.3)} & 53.4\,{\scriptsize($-$1.1)} & 81.1\,{\scriptsize(+0.0)} & 69.4\,{\scriptsize($-$0.9)} & 12 & 1.06 \\
Token pruning & One-shot drop & drop 70\% & 69.4\,{\scriptsize($-$1.6)} & 54.2\,{\scriptsize($-$0.3)} & 80.5\,{\scriptsize($-$0.6)} & 69.3\,{\scriptsize($-$1.0)} & \textbf{32} & \textbf{1.27} \\
Layer skip & Trained gate\,{\scriptsize(App.~\ref{sec:trained-arms})} & skip 14\% & 70.0\,{\scriptsize($-$1.0)} & 52.6\,{\scriptsize($-$1.9)} & 81.3\,{\scriptsize(+0.2)} & 69.5\,{\scriptsize($-$0.8)} & 7.7 & 1.17 \\
Early exit & Multi-exit & exit at 36/40 & 69.3\,{\scriptsize($-$1.7)} & 52.8\,{\scriptsize($-$1.7)} & 79.6\,{\scriptsize($-$1.5)} & 68.7\,{\scriptsize($-$1.6)} & 5.4 & 1.12 \\
\midrule
\multicolumn{9}{@{}l}{\textit{Inference-only methods}}\\
\midrule
\multirow{3}{*}{Adaptive top-$k$} & Uniform top-$k$ & 50\% skip & 70.5\,{\scriptsize($-$0.5)} & 53.6\,{\scriptsize($-$0.9)} & 80.9\,{\scriptsize($-$0.2)} & 69.8\,{\scriptsize($-$0.5)} & 11.6 & 1.06 \\
 & MoDES (DMT)~\citep{huang2025modes} & 59\% skip & 70.4\,{\scriptsize($-$0.6)} & 53.6\,{\scriptsize($-$0.9)} & 80.9\,{\scriptsize($-$0.2)} & 69.8\,{\scriptsize($-$0.5)} & 13.8 & 1.04 \\
 & Random & 50\% skip & 63.2\,{\scriptsize($-$7.8)} & 46.4\,{\scriptsize($-$8.1)} & 77.5\,{\scriptsize($-$3.6)} & 63.7\,{\scriptsize($-$6.6)} & 11.6 & 1.05 \\
\cmidrule(l){1-9}
\multirow{4}{*}{Expert pruning} & REAP~\citep{lasby2025reap} & \multirow{4}{*}{$E{=}128$} & 70.3\,{\scriptsize($-$0.7)} & 53.8\,{\scriptsize($-$0.7)} & 80.1\,{\scriptsize($-$1.0)} & 69.5\,{\scriptsize($-$0.8)} & \multirow{4}{*}{0.3} & \multirow{4}{*}{1.00} \\
 & MSAN~\citep{liu2026scoreexperts} &  & 69.9\,{\scriptsize($-$1.1)} & 54.0\,{\scriptsize($-$0.5)} & 80.2\,{\scriptsize($-$0.9)} & 69.4\,{\scriptsize($-$0.9)} &  &  \\
 & MAN~\citep{liu2026scoreexperts} &  & 70.1\,{\scriptsize($-$0.9)} & 53.6\,{\scriptsize($-$0.9)} & 80.3\,{\scriptsize($-$0.8)} & 69.4\,{\scriptsize($-$0.9)} &  &  \\
 & Router-prob &  & 68.8\,{\scriptsize($-$2.2)} & 52.7\,{\scriptsize($-$1.8)} & 80.4\,{\scriptsize($-$0.7)} & 68.7\,{\scriptsize($-$1.6)} &  &  \\
\cmidrule(l){1-9}
\multirow{2}{*}{Token pruning} & GDN-$\beta$ & \multirow{2}{*}{keep 50\%} & 69.7\,{\scriptsize($-$1.3)} & 53.5\,{\scriptsize($-$1.0)} & 80.2\,{\scriptsize($-$0.9)} & 69.2\,{\scriptsize($-$1.1)} & \multirow{2}{*}{22.8} & 1.21 \\
 & Random &  & 69.0\,{\scriptsize($-$2.0)} & 52.8\,{\scriptsize($-$1.7)} & 79.4\,{\scriptsize($-$1.7)} & 68.4\,{\scriptsize($-$1.9)} &  & 1.19 \\
\bottomrule\end{tabular}}
\end{table}

\begin{figure}[t]\centering
  \includegraphics[width=\textwidth]{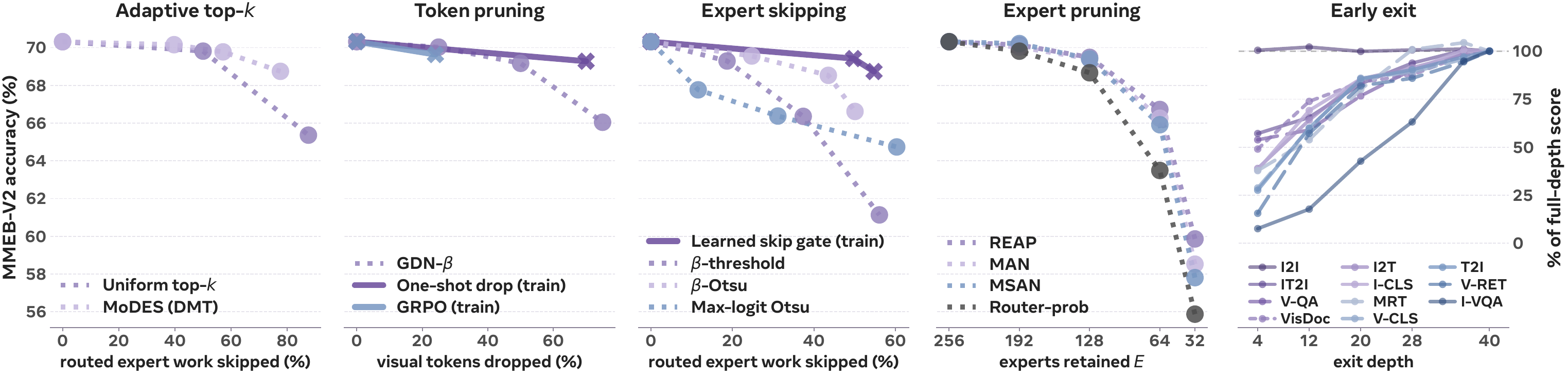}
  \vspace{-15pt}
  \caption{\footnotesize Compute-accuracy curves per method. More details
  are reported in Tab.~\ref{tab:inference} and App.~\ref{sec:additional-efficiency}.}
  \label{fig:budget-accuracy}
  \vspace{-15pt}
\end{figure}


\textbf{Adaptive computation yields strong accuracy--compute trade-offs for MoE-based UME.} Overall, visual token pruning reduces execution cost notably and static expert pruning halves the expert pool, both at minimal accuracy loss (Tab.~\ref{tab:inference}), confirming substantial token-level and parameter-level redundancy in MoE-UME.
\vspace{-2pt}

\textbf{Token pruning has the strongest end-to-end efficiency; expert-level actions give the weakest.} We observe token pruning to be the most effective adaptive computation approach, converting significant FLOP savings directly into throughput gains at minimal accuracy cost. By contrast, adaptive top-$k$ methods yield minor wall-clock speedups.

\textbf{Inference-only adaptive approaches are competitive with training-based methods.} Inference-only MoDES outperforms trained adaptive top-$k$~\citep{zeng2024adamoe}, while our GDN-$\beta$-based pruning rivals trained token pruning gates. In general, we observe training-based method to perform slightly better mainly under extreme pruning regimes (Fig.~\ref{fig:budget-accuracy}).


\begin{wrapfigure}{r}{0.5\textwidth}
\centering
\vspace{-21pt}
\includegraphics[width=\linewidth]{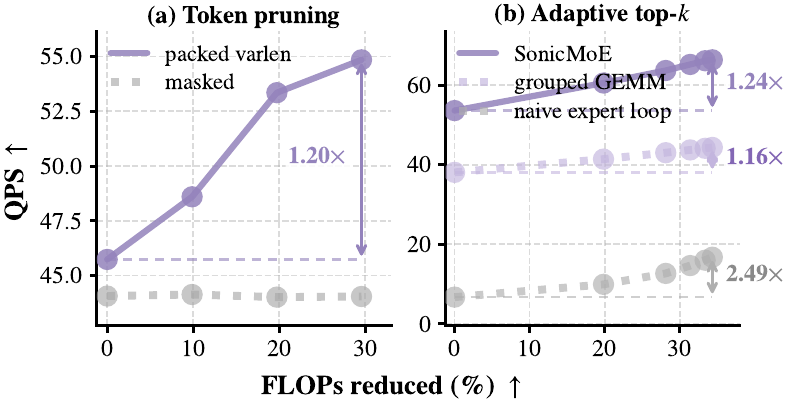}
\vspace{-20pt}
\captionsetup{font=footnotesize}
\caption{Transferring FLOPs saving into wall-clock speedup under baseline (grey) and our implementations.}
\label{fig:wallclock}
\vspace{-25pt}
\end{wrapfigure}
\textbf{Exit depths vary widely across tasks.} Fig.~\ref{fig:budget-accuracy} shows results for early exit. We can observe (1) Supervising intermediate exit depths has negligible impact on full-depth accuracy; (2) Exit depth depends heavily on task complexity: symmetric I2I retrieval plateaus after 4 of 40 layers, which confirms our hypothesis that the LLM layers can bring significant redundancy for simple retrievals. On the other hand, complex, text-guided retrievals require substantially deeper layers.

\textbf{Our implementations effectively convert FLOP savings to wall-clock speedups.} As shown in Fig.~\ref{fig:wallclock}, packed varlen and SonicMoE convert 56--82\% of FLOP reductions into runtime gains, far outperforming unmodified grouped GEMM (41\%) and masking (0\%). Baseline implementations fail to leverage adaptive top-$k$ efficiently: grouped GEMM is tile-bound and saturates past $k \approx 4$, while the naive HuggingFace loop remains $5.7$--$8.0\times$ slower overall despite high relative scaling.


\begin{figure}[htbp]\centering
\vspace{-7pt}
  \includegraphics[width=\textwidth]{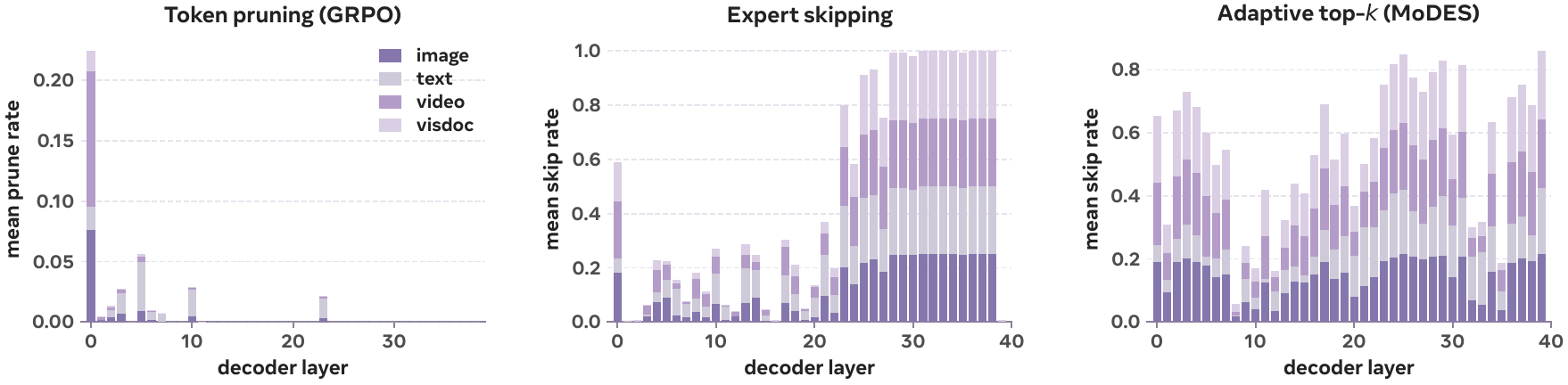}
  \vspace{-15pt}
  \caption{Depth-wise adaptive compute allocation across modalities.}
  \vspace{-10pt}
  \label{fig:depth-allocation}
\end{figure}


\textbf{Depth-wise adaptive compute allocation across modalities.} Layer-wise allocation patterns (Fig.~\ref{fig:depth-allocation}) reveal distinct modality behaviors: \textbf{Token pruning} acts early, immediately removing redundant image and video tokens while preserving text queries and document layouts for early contextualization. \textbf{Expert skipping} exhibits higher early-layer skip rates for text before unifying across modalities in deep layers and dropping to near-zero at the final layer for output embedding construction. Conversely, \textbf{adaptive top-$k$} allocates more experts to text in earlier layers.

\section{Conclusion}
\label{sec:conclusion}


We presented \method{}, the first sparse MoE-based UME model. This approach gains over 10 accuracy points at near-constant active compute, establishing a new SOTA on MMEB-V2 and MRMR among public-data models. Furthermore, we provide the systematic study over adaptive computation methods for MoE-based embedding models, where we show that MoE embedders can reduce compute budget significantly with under a one-point accuracy loss.

\section*{Reproducibility Statement}

The settings behind each reported number are documented in the paper. Backbone
configurations, the training data mixture with its sampling weights, and the
optimization settings for both \method{}-A3B and \method{}-A10B are given in
\S\ref{sec:setup} and App.~\ref{sec:implementation-details};
App.~\ref{sec:routing-objectives} formalizes the router adaptation variants
compared in Tab.~\ref{tab:recipe}. For adaptive computation, the per-method
training setups appear in App.~\ref{sec:trained-arms}, and the inference-only
procedures, with their calibration, in App.~\ref{sec:inference-details}. The
efficiency claims depend on how compute is counted, so
App.~\ref{sec:measurement-details} states the FLOP accounting, the hardware and
batching conditions under which throughput was measured, and the construction of
Fig.~\ref{fig:flops-acc}. MMEB-V2 and MRMR are both public and we follow their
official evaluation protocols; App.~\ref{sec:per-task} lists our per-task scores
so that any individual entry can be checked against them.

\subsection*{AI use statement}

Anthropic Claude Code (Opus 4.8 and Opus 5) was used extensively for drafting code, aggregating results, and drafting manuscript. The authors verified experiment artifacts, numerical
claims, and citations and take responsibility for the content.

\newpage

\bibliographystyle{assets/plainnat}
\bibliography{references}

\clearpage
\newpage

\newpage
\section*{Appendix}
\startcontents[appendix]
\printcontents[appendix]{}{1}{\setcounter{tocdepth}{2}}
\vspace{0.6em}

\newpage

\section{Related Work}
\label{sec:related-work}

\textbf{Universal Multimodal Embeddings.}
Recent advances in multimodal representation learning adapt vision-language models into general-purpose encoders evaluated on broad benchmark suites such as MMEB and MMEB-V2~\citep{jiang2025vlm2vec,meng2025vlm2vecv2}. Prior efforts to improve universal embedding performance primarily focus on data supervision, including hard-negative mining~\citep{thirukovalluru2025b3,lin2025mmembed,gu2025unimev2} and multi-stage reranker distillation~\citep{li2026qwen3vlembedding,cui2026reasontocontrast}. Another emerging direction leverages test-time reasoning, optimizing autoregressive thought generation prior to embedding emission~\citep{lan2025umer1,jiang2026embedrl,cheng2026tteflash,zhang2026thinkneeded}. In contrast to data-centric scaling or test-time generation overhead, we examine the complementary axis of encoder parameter capacity, evaluating how native sparse expert scaling improves contrastive representations. 

Our work also differs fundamentally from existing MoE-based embedding approaches. TSEmbed~\citep{wu2026tsembed} employs fixed task- or modality-wise LoRA routing, which caps expert capacity by the task taxonomy. We instead adapt native token-level sparse MoE backbones, allowing capacity to scale with expert count via dynamic, learned routing. Furthermore, unlike MoEE~\citep{li2024moeembedding}, which extracts off-the-shelf text embeddings directly from frozen router probabilities, we fine-tune sparse multimodal architectures to pool high-quality representation states.

\textbf{Sparse Mixture-of-Experts Backbones.}
Sparsely gated MoE architectures scale model capacity while holding per-token compute constant~\citep{shazeer2017outrageously,fedus2022switch}. Modern foundation architectures increasingly adopt fine-grained sparse routing to expand representation power, as demonstrated in Mixtral~\citep{jiang2024mixtralexperts}, DeepSeek-V3~\citep{deepseekai2024deepseekv3}, OLMoE~\citep{muennighoff2024olmoe}, Kimi K3~\citep{kimiteam2026kimik3openfrontier}, Qwen3.5~\citep{qwenteam2026qwen35}, and Aria~\citep{li2025aria}. However, these routing mechanisms are pre-trained exclusively under next-token prediction. We study how these pre-trained sparse routing distributions transfer when adapted contrastively to produce fixed-length vector representations.

\textbf{Efficient Embeddings and Adaptive Computation.}
Adjustable-cost embedding methods have been studied via Matryoshka dimension truncation~\citep{kusupati2022matryoshka,li2024twodmse}, variable late-interaction vectors~\citep{faysse2025colpali,xiao2025metaembed}, and visual patch reduction~\citep{ma2025storageefficientvdr}. However, these works focus on reducing retrieval and storage cost, not the active computation for calculating the embedding. In parallel, adaptive computation frameworks reduce inference latency via dynamic depth skipping~\citep{banino2021pondernet,raposo2024mixtureofdepths}, visual token pruning~\citep{chen2024fastv,huang2025dynamicllava}, and adaptive top-$k$ routing~\citep{huang2025modes}, but these works mostly focus on visual classification and next token prediction (text generation) tasks, are mostly explored separately, and have not been thoroughly tested on embedding task. We unify these paradigms by systematically comparing token-, expert-, and layer-level execution controls within a single MoE multimodal embedder, identifying which mechanisms lead to best accuracy-efficiency tradeoff, and which best translate theoretical FLOP reductions into deployed wall-clock speedups.

\section{Limitations and future work}
\label{sec:limitations}

While \method{} achieves SOTA performance over models trained with public MMEB-V2 training data, it's currently lagging behind models trained with external data. Also, it currently does not support audio as input modality. Our future work include training \method{} under larger scale of data, and to include more input modalities (\textit{e.g.} audio) and more diverse tasks.

\section{Implementation Details}
\label{sec:implementation-details}

\subsection{Model Architecture and Experimental Setup}\textbf{Backbone Configurations.} We evaluate two primary variants built upon native sparse MLLM backbones: \method{}-A3B (3.1B active parameters) and \method{}-A10B (10.2B active parameters). Both variants employ $E = 256$ routed experts per layer with top-8 routing ($k = 8$) and a single shared expert. \method{}-A3B uses a hidden dimension of $d = 2048$, 40 decoder layers (30 Gated-DeltaNet and 10 full-attention blocks), and an expert intermediate dimension of $d_{\text{ffn}} = 512$. \method{}-A10B expands model capacity with $d = 3072$, 48 decoder layers, and $d_{\text{ffn}} = 1024$.

\textbf{Training Data.} Models are fine-tuned on a multi-task mixture comprising 24 sampling streams across image, video, and visual-document retrieval. This includes 17 image-text tasks from MMEB~\citep{jiang2025vlm2vec}, 4 visual-document streams from ViDoRe and VisRAG, and 3 video retrieval streams derived from LLaVA-Hound. Following our standard protocol, we exclude generated chain-of-thought sequences from training targets. Complete dataset sampling weights are detailed in Tab.~\ref{tab:training-subsets}.

\textbf{Training setups.} We train for two epochs using BF16 precision and the AdamW optimizer ($\beta_1 = 0.9, \beta_2 = 0.98, \text{weight decay} = 0.01$). The learning rate warms up to $10^{-4}$ over 10 steps, followed by a cosine decay to $10^{-5}$. Training uses a global batch size of 512 across 64 H200 GPUs with a maximum sequence length of 8,192 tokens. To enable parameter-efficient adaptation, we apply LoRA ($r = 64, \alpha = 128$) to the vision encoder, attention projections, and expert feed-forward layers, yielding 73.6M trainable parameters for \method{}-A3B and 196.6M for \method{}-A10B. 

\begin{table}[ht]
\centering
\caption{\textbf{Training subsets.}  Weights are relative and do not sum to one.}
\label{tab:training-subsets}
\footnotesize
\setlength{\tabcolsep}{4pt}
\begin{tabular}{lr@{\hspace{16pt}}lr@{\hspace{16pt}}lr}
\toprule
Subset & Weight & Subset & Weight & Subset & Weight \\
\midrule
  A-OKVQA & 0.26 & ChartQA & 0.35 & CIRR & 0.43 \\
  DocVQA & 0.84 & ImageNet-1K & 2.25 & InfographicsVQA & 0.31 \\
  MSCOCO i2t & 2.58 & MSCOCO image pairs & 3.78 & MSCOCO t2i & 2.32 \\
  NIGHTS & 0.23 & OK-VQA & 0.25 & SUN397 & 0.22 \\
  VisDial & 3.75 & Visual7W & 1.68 & VisualNews i2t & 2.91 \\
  VisualNews t2i & 3.49 & WebQA & 0.23 & ViDoRe subset 1 & 5.00 \\
  ViDoRe subset 2 & 5.00 & VisRAG subset 1 & 6.00 & VisRAG subset 2 & 6.00 \\
  Video caption & 5.27 & Video QA & 4.38 & Video retrieval & 5.76 \\
\bottomrule
\end{tabular}
\end{table}

\subsection{Router Adaptation Objectives and Priors}
\label{sec:routing-objectives}

We detail the formulations for the router adaptation variants evaluated in Tab.~\ref{tab:recipe}.

\textbf{Load-Balancing and $z$-Loss.} To evaluate whether standard pre-training regularizers aid contrastive adaptation, we include a joint auxiliary loss $\mathcal{L}_{\text{aux}} = \lambda_{\text{lb}} \mathcal{L}_{\text{lb}} + \lambda_z \mathcal{L}_z$. The load-balancing objective $\mathcal{L}_{\text{lb}} = E \sum_{j=1}^E f_j P_j$ encourages uniform expert allocation~\citep{fedus2022switch}, where $f_j$ is the fraction of tokens routed to expert $j$ and $P_j$ is the mean router probability assigned to expert $j$. The router $z$-loss $\mathcal{L}_z = \frac{1}{B} \sum_{b=1}^B (\log \sum_{j=1}^E e^{g_{bj}})^2$ penalizes large pre-softmax router logits $g$ to stabilize routing dynamics~\citep{zoph2022stmoe}. We set weighting coefficients to $\lambda_{\text{lb}} = 0.01$ and $\lambda_z = 0.001$.

\textbf{Conflict-Aware Routing.} To mitigate multi-task expert interference, this objective forces task pairs with opposing router gradients onto distinct experts. For task $t$ at layer $\ell$, we maintain exponential moving averages of its expert assignment distribution $\bar{\pi}^{(\ell)}_t \in \mathbb{R}^E$ and router gradient $g^{(\ell)}_t$. The gradient conflict between current task $s$ and candidate task $t$ is defined as $C^{(\ell)}_{st} = \operatorname{clamp}(-\cos(g^{(\ell)}_s, g^{(\ell)}_t), 0, 1)$. The objective maximizes the conflict-weighted Jensen--Shannon (JS) divergence between routing distributions across layers:
\begin{equation}
\mathcal{L}_{\mathrm{conf}} = -\frac{1}{|\mathcal{P}|} \sum_{t \in \mathcal{P}} \frac{1}{L} \sum_{\ell=1}^L C^{(\ell)}_{st} \, \operatorname{JS}\left(\bar{\pi}^{(\ell)}_s \parallel \bar{\pi}^{(\ell)}_t\right),
\end{equation}
where $\mathcal{P}$ is the set of tasks with non-zero conflict relative to task $s$, and $\operatorname{JS}(p \parallel q) = \frac{1}{2} \operatorname{KL}(p \parallel m) + \frac{1}{2} \operatorname{KL}(q \parallel m)$ for mixture distribution $m = \frac{1}{2}(p + q)$. This pushes conflicting task pairs toward disjoint expert subsets while exerting no penalty on cooperative or orthogonal pairs.

\textbf{Task-ID and Instruction Priors.} We test conditioning router assignments on task information by injecting an additive bias vector $b^{(\ell)} \in \mathbb{R}^E$ into the router logits: $g^{(\ell)} = W_r^{(\ell)} h + b^{(\ell)}$. We test two setups: \emph{(1) task-ID prior}: $b^{(\ell)}$ is obtained from a learned task embedding table with hidden size 256 and dropout 0.3. \emph{(2) instruction prior}, $b^{(\ell)}$ is projected from the mean pooling of instruction tokens at layer $\ell$, updating the routing bias dynamically across network depth.

\section{Measurement Protocol for FLOPs and Throughput}
\label{sec:measurement-details}

\subsection{Counting Canonical FLOPs}

Logical FLOP counts are computed using the same query/target (whichever side contains multimodal input) subset per task across all 78 MMEB-V2 tasks.  As the regular $\operatorname{FlopsCounter}$ does not capture fused kernels such as FlashAttention~\citep{flashattention2} and linear attention~\citep{yang2024fla}, their FLOPS are added analytically. Grouped expert GEMMs are computed based on physically executed row volumes to prevent double-counting pruned expert allocations.

\subsection{Accuracy--Throughput Figure Construction}
\label{sec:fig1-details}

In this section we provide details for constructing Fig.~\ref{fig:flops-acc}.

\textbf{\method{} Configurations.} The expert-scaling curve for \method{}-A3B evaluates the performance with different number of total experts $E \in \{32, 64, 128, 192, 256\}$, constructed by pruning the full $E=256$ model using REAP~\citep{lasby2025reap} under the reconstruction-error criterion (Eq.~\ref{eq:prune-scores}). \method{}-A10B represents the full $E = 256$ backbone evaluated under the identical protocol. The single adaptive compute marker corresponds to the GRPO-trained policy (App.~\ref{sec:controller-details}).

\textbf{Baseline Models.} Autoregressive think-then-embed systems, such as Embed-RL~\citep{jiang2026embedrl} and UME-R1~\citep{lan2025umer1}, are evaluated with reasoning generation both enabled and disabled to highlight their accuracy-throughput tradeoff via reasoning. PLUME~\citep{he2026plume} scales test-time compute across latent reasoning steps $K \in \{4, 6, 8\}$, under which throughput drops from 4.5 to 3.2 queries/sec. VLM2Vec-V2~\citep{meng2025vlm2vecv2} serve as single-vector dense references.

\subsection{Converting FLOP Savings into Wall-Clock Speedups}
\label{sec:wallclock-details}

This section details the empirical latency measurements supporting Fig.~\ref{fig:wallclock}.

A reduction in logical computation is not a reduction in QPS, and the gap
decides which actions are worth having.  However, a large body of adaptive computation
works focus the former and ignore the latter~\citep{zeng2024adamoe,huang2024dynamicrouting}. In this work, we address this gap by providing implementation for effectively converting the computation saving into actual batched inference acceleration. We then conduct detailed benchmarking for both computation and wall-clock savings.

For token pruning-based actions we adopt a \emph{packed variable-length} approach: at
each layer the surviving tokens are gathered into contiguous segments with updated
cumulative sequence lengths, and a segment index keeps the attention operations from mixing
adjacent samples. This procedure allows for batched adaptive computation where token numbers vary across layers, which is a nontrivial implementation compared to one-shot pruning~\citep{huang2025dynamicllava,shang2024prumerge}.

For MoE-specific methods, many
report efficiency without an inference wall-clock
measurement~\citep{zeng2024adamoe,huang2024dynamicrouting,li2023adaptivegating},
do not supply the implementation that would realize the
saving~\citep{huang2025modes}, or rely on a naive implementation that loops over
experts~\citep{lu2024notallexperts}, which we find to be $5.7$--$8.0\times$
slower than a fused kernel, so any speedup measured against it is overstated. In this work, we run the wall-clock benchmarks through tile-aware grouped-GEMM
kernels~\citep{guo2025sonicmoe} alongside the batched grouped
GEMM~\citep{gale2022megablocks}, which exposes where per-token expert
sparsity stops gaining wall-clock speedup from FLOPS saving.


\textbf{Sequential Expert Looping.} Sequential expert execution used in some existing MoE-efficiency methods~\citep{lu2024notallexperts} incurs heavy kernel launch overhead, running $5.7\text{--}8.0\times$ slower than fused grouped GEMMs ($1189.5\text{ ms}$ vs.\ $210.2\text{ ms}$ for grouped GEMM at $B=8, S=1024$; Tab.~\ref{tab:backend-matrix}). Although sequential loops exhibit high sensitivity to expert sparsity (reducing latency by $32.3\text{--}59.9\%$), this relative gain stems from high baseline launch overheads rather than efficient execution. 

\begin{table}[ht]
\centering
\caption{Different throughput and FLOPS corresponding to different expert skip rate, for different MoE kernels. Throughput is queries per second on one H200.}
\label{tab:backend-matrix}
\small
\setlength{\tabcolsep}{5pt}
\begin{tabular}{rrr rrr}
\toprule
\multicolumn{3}{c}{\textbf{Policy}} & \multicolumn{3}{c}{\textbf{Throughput (QPS)}} \\
\cmidrule(lr){1-3}\cmidrule(lr){4-6}
Mean $k$ & Skip (\%) & FLOPs $\downarrow$\% & Naive loop & Grouped GEMM & SonicMoE \\
\midrule
8.00 & 0.0 & 0.0 & 6.7 & 38.1 & 53.6 \\
4.02 & 49.8 & 19.8 & 9.9 & 41.4 & 60.5 \\
2.35 & 70.6 & 28.1 & 12.7 & 43.0 & 63.7 \\
1.70 & 78.8 & 31.4 & 14.7 & 43.7 & 65.2 \\
1.30 & 83.8 & 33.4 & 16.0 & 44.1 & 66.1 \\
1.10 & 86.2 & 34.3 & 16.8 & 44.3 & 66.4 \\
\bottomrule
\end{tabular}
\end{table}

\textbf{Tile-Bound Constraints in Grouped GEMMs.} Grouped GEMM
kernels~\citep{gale2022megablocks} process each expert's tokens in tiles of at
least $M=128$ rows, and a partly filled tile costs mostly the same time as a full one.  An
expert holding $n_j$ routed rows therefore requires $\lceil n_j/M \rceil$ tiles, and skipping tokens does not produce actual wall-clock saving until it empties a whole tile.  Whether it does depends on how many rows an expert holds to begin with.  Under
top-8 routing a batch of $T$ tokens produces $8T$ routed token--expert pairs,
which spread over $E=256$ experts give $n_j \approx 8T/E = T/32$ rows apiece
when the router is balanced, so the count grows with sequence length.

At $T=4{,}096$ tokens that is roughly 128 rows per expert.  A $52.9\%$ skip policy cuts that to about 50, but both amounts occupy a single tile, so the kernel executes the work it would have executed anyway while the model still spends time on the additional skip gates, which results in a $0.91\times$ slowdown.  At $T=8{,}192$ each expert instead holds around 256 rows, which is two tiles.  There the same kind of policy at $69.6\%$ skip cuts the count to about 78 rows, one tile ($\lceil 256/128 \rceil = 2 \to \lceil 78/128 \rceil = 1$), and removing that second tile is a genuine saving of $1.11\times$. This suggests that the wall-clock speedup for adaptive top-$k$ or expert skipping also depends on the sequence length or the batch size.

\textbf{Backend Customization and Speedup Saturation.} Hardware speedups are primarily driven by backend kernel design (Fig.~\ref{fig:wallclock}b). Replacing standard grouped GEMM with SonicMoE~\citep{guo2025sonicmoe} kernel yields an $1.40\times$ speedup at fixed $k=8$ ($210.2\text{ ms}$ vs.\ $149.2\text{ ms}$ at $S=8192$). Within SonicMoE, lowering active expert width from $k=8$ to $k \approx 4$ provides an additional $1.11\text{--}1.24\times$ speedup. Further reduction to $k \approx 1.1$ yields diminishing returns ($1.17\times$ total for grouped GEMM, $1.74\times$ for SonicMoE; Tab.~\ref{tab:backend-matrix}) as execution becomes dominated by non-routed operations, including self-attention, shared experts, and token dispatch.

\begin{wraptable}{r}{0.54\textwidth}
\vspace{-1.1\baselineskip}
\centering
\caption{Token pruning under masking and physical compaction.}
\label{tab:token-compaction}
\small
\setlength{\tabcolsep}{5.5pt}
\begin{tabular}{r rr rrr}
\toprule
 & \multicolumn{2}{c}{\textbf{Masked}} & \multicolumn{3}{c}{\textbf{Packed varlen}} \\
\cmidrule(lr){2-3}\cmidrule(lr){4-6}
Keep & QPS & Speedup & FLOPs $\downarrow$\% & QPS & Speedup \\
\midrule
1.00 & 45.7 & 1.061 & 0.0 & 45.7 & 1.024 \\
0.75 & 44.1 & 1.025 & 9.9 & 48.6 & 1.088 \\
0.50 & 44.0 & 1.023 & 19.7 & 53.4 & 1.194 \\
0.25 & 44.0 & 1.023 & 29.6 & 54.9 & 1.228 \\
\bottomrule
\end{tabular}
\vspace{-1.0\baselineskip}
\end{wraptable}

\textbf{Physical Sequence Compaction vs.\ Spatial Masking.} Token pruning requires physical sequence compaction to achieve wall-clock latency reductions. Spatial masking preserves activation tensor shapes ($[B, S, H]$), resulting in a flat $1.02\times$ execution speedup across all pruning budgets (Fig.~\ref{fig:wallclock}a). In contrast, physically packing remaining sequence tokens post-pruning via FlashAttention-2 varlen function~\citep{flashattention2} shortens activation tensors across downstream layers, yielding speedups of $1.088\times$, $1.194\times$, and $1.228\times$ at $75\%$, $50\%$, and $25\%$ visual token keep ratios (Tab.~\ref{tab:token-compaction}).

\section{Additional Studies on MoE Design}
\label{sec:design-studies}

This section expands the design and capacity ablations of
\S\ref{sec:scaling}.

\FloatBarrier

\subsection{Dense-to-MoE Upcycling}
\label{sec:upcycling}

To evaluate whether expert capacity can be synthesized during downstream fine-tuning, we upcycle a dense 2B multimodal encoder into a sparse MoE architecture by varying structural granularities and router initialization strategies.

\textbf{Structural Granularity.} A virtual group count $G$ determines how the dense feed-forward network (FFN) is partitioned. For coarse upcycling ($G = 1$), each expert is a full replica of the original FFN ($d_{\text{expert}} = d_{\text{ffn}}$), where top-1 routing reproduces the dense baseline's FLOP footprint. For fine-grained upcycling ($G > 1$), the FFN is sliced along its intermediate dimension into $G$ narrower experts ($d_{\text{expert}} = d_{\text{ffn}}/G$), requiring $k = G$ active experts per token to match the capacity of the dense layer.

\textbf{Router Initialization.} Because the router gate is newly introduced, its initialization controls initial expert activation patterns. We test three initialization methods: (1) \emph{random} initialization ($\mathcal{N}(0, 0.02)$) breaks symmetry between identical experts; (2) \emph{tiled} initialization replicates the dense gate row across all experts; and (3) \emph{zero$+$bias} sets initial pre-softmax logits to zero with a positive bias, ensuring every FFN slice is selected from the first step so the layer matches dense output prior to adaptation.

\begin{table}[ht]
\centering
\caption{\textbf{Dense-to-MoE upcycling on the 2B encoder.}  Image, Video and VisDoc are the 36, 18 and 24 task averages; Overall is over all 78.}
\label{tab:upcycling}
\small
\setlength{\tabcolsep}{5pt}
\begin{tabular}{rrrlr rrrr}
\toprule
$E$ & $G$ & $k$ & Init & Act. & Image & Video & VisDoc & Overall \\
\midrule
\multicolumn{5}{@{}l}{Dense reference} & 62.5 & 42.6 & 77.2 & 62.41 \\
\midrule
4 & 1 & 2 & tiled & $2\times$ & 62.6 & 45.3 & 77.3 & \textbf{63.12} \\
4 & 1 & 1 & random & $1\times$ & 63.8 & 40.0 & 77.2 & 62.44 \\
32 & 1 & 1 & random & $1\times$ & 61.0 & 38.8 & 75.2 & 60.26 \\
32 & 4 & 4 & zero+bias & $1\times$ & 60.3 & 40.0 & 73.4 & 59.64 \\
8 & 4 & 4 & zero+bias & $1\times$ & 59.5 & 40.4 & 73.8 & 59.50 \\
8 & 4 & 4 & tiled & $1\times$ & 59.9 & 34.9 & 72.1 & 57.88 \\
\bottomrule
\end{tabular}
\end{table}

\textbf{Results.} As shown in Tab.~\ref{tab:upcycling}, upcycling fails to improve representation quality over the dense baseline under same active compute. Coarse top-1 upcycling ($E=4, G=1$) performs comparably to the dense reference (62.44 vs.\ 62.41), while increasing expert count ($E=32$) causes performance to drop to 60.26. Fine-grained partitioning ($G=4$) similarly reduces overall accuracy below 60.0. Outperforming the dense baseline requires doubling the active compute per token ($2\times$, top-2 routing), reaching 63.12. These results confirm that synthesizing expert capacity during contrastive fine-tuning is ineffective, highlighting the necessity of native MoE backbones pre-trained at scale.

\subsection{Capacity Ablation on Specialist and Generalist Models}
\label{sec:capacity-ablation}

\textbf{Experimental Protocol.}
To evaluate how model capacity impacts multi-task adaptation, we evaluate multi-task generalists against task-group specialists across three backbones: Qwen3.5-2B (dense, 2B active), Qwen3.5-9B (dense, 9B active), and Qwen3.5-35B-A3B (MoE, 3.1B active). Specialist models are fine-tuned independently on individual task groups (Image Classification and Image Retrieval), whereas generalists are fine-tuned on the full dataset mixture. We measure multi-task capacity degradation using the generalist-specialist performance gap. All configurations share identical training hyperparameters (global batch size 256, learning rate $5\times10^{-5}$, LoRA rank 64, $\alpha=128$, $\texttt{<emb>}$ pooling, 1 epoch per setting).

\begin{figure}[htbp]\centering
  \includegraphics[width=0.62\textwidth]{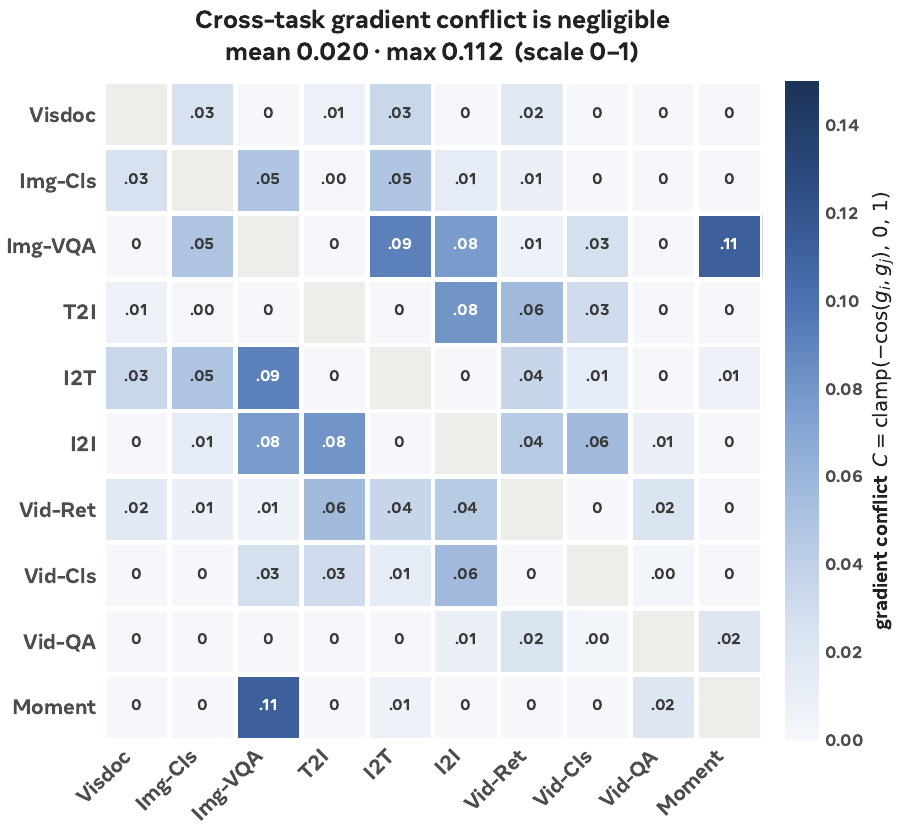}
  \caption{Pairwise conflict $C=\mathrm{clamp}(-\cos(g_i,g_j),0,1)$ on the \texttt{<emb>} gradient.}
  \label{fig:task-conflict}
\end{figure}

\textbf{Minimal Cross-Task Gradient Conflict.}
A primary hypothesis for multi-task performance degradation is destructive gradient interference across task streams. To test this, we compute the pairwise cosine conflict $C = \operatorname{clamp}(-\cos(g_i, g_j), 0, 1)$ across the ten MMEB-V2 categories, where $g$ is the gradient of $\mathcal L_{\mathrm{emb}}$ (Eq.~\ref{eq:infonce}) with respect to the pooled \texttt{<emb>} embeddings $q$ and $t$, averaged within a category. As shown in Fig.~\ref{fig:task-conflict}, the conflict is negligible (mean $0.020$, max $0.112$). The router-parameter gradients that $\mathcal{L}_{\mathrm{conf}}$ acts on (App.~\ref{sec:routing-objectives}) conflict even less at the same checkpoint: mean $0.003$ and max $0.035$ over the nine training task types. Because categories do not pull the \texttt{<emb>} embedding in opposing directions, generalist degradation stems primarily from total capacity constraints rather than inter-task gradient competition. Expanding expert capacity directly addresses this limitation~\ref{fig:capacity-ablation}.

\begin{table}[htbp]\centering
\caption{\textbf{Specialist vs.\ generalist by backbone.}  Group scores are unweighted averages over
the group's tasks; \textbf{Mean} is the flat mean of the two group scores and
is what Fig.~\ref{fig:capacity-ablation} plots.}
\label{tab:capacity-ablation}
\footnotesize\setlength{\tabcolsep}{5pt}\renewcommand{\arraystretch}{1.1}
\begin{tabular}{llccc}
\toprule
\textbf{Backbone} & \textbf{Arm} & \textbf{Image-CLS} & \textbf{Image-Ret.} & \textbf{Mean}\\
 & & \scriptsize(10 tasks) & \scriptsize(16 tasks) & \\
\midrule
\multirow{3}{*}{Qwen3.5-2B \scriptsize(dense, 2B act.)}
 & specialist & 60.69 & 73.06 & 66.88\\
 & generalist & 56.53 & 71.17 & 63.85\\
 & $\Delta$   & $-4.16$ & $-1.89$ & $-3.02$\\
\midrule
\multirow{3}{*}{Qwen3.5-9B \scriptsize(dense, 9B act.)}
 & specialist & 63.50 & 77.49 & 70.50\\
 & generalist & 61.34 & 75.81 & 68.58\\
 & $\Delta$   & $-2.16$ & $-1.68$ & $-1.92$\\
\midrule
\multirow{3}{*}{Qwen3.5-35B-A3B \scriptsize(MoE, 3.1B act.)}
 & specialist & 59.73 & 74.04 & 66.89\\
 & generalist & 58.97 & 72.54 & 65.75\\
 & $\Delta$   & $\mathbf{-0.76}$ & $\mathbf{-1.50}$ & $\mathbf{-1.13}$\\
\bottomrule
\end{tabular}
\end{table}

\clearpage
\subsection{Expert Routing Behavior Across Modality and Task}
\label{sec:expert-routing}

To analyze expert routing patterns across modalities and task domains, we profile top-8 routing assignments across all 40 MoE layers using 10 MMEB-V2 categories covering all modalities and tasks, with 16 samples per category. The routing divergences are measured using Jensen--Shannon (JS) distance over the 256-expert distribution.

\begin{wrapfigure}{r}{0.60\textwidth}
  \centering
  \includegraphics[width=\linewidth]{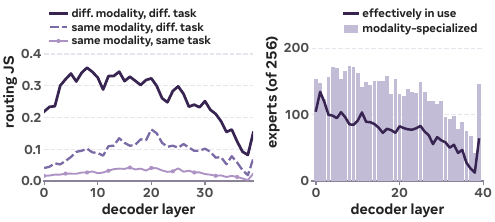}
  \caption{\textbf{Expert routing separates input types, and the separation
  fades with depth.}  (a) Per-layer JS distance between two of the 20 subsets,
  each summarized by its 256-expert histogram, averaged over three disjoint
  sets of pairs.  Crossing a task boundary at fixed input type stays near the
  same-task curve at every layer, while crossing an input type runs about three
  times higher.  (b) How many of the 256 experts have specialized toward a
  single modality ($|\pi_j| > 0.5$), and how many the layer keeps occupied
  ($\exp(H)$ of the routing entropy).}
  \label{fig:expert-routing}
\end{wrapfigure}

\textbf{Modality Preference Over Task.} Grouping subsets by input type (photo, document, video, text) shows that expert routing depends far more on input modality than task taxonomy (Fig.~\ref{fig:expert-routing}a). Photos and videos draw on heavily overlapping expert populations, whereas document pages and text branch into distinct pathways. Notably, this modality specialization emerges naturally during contrastive adaptation without needing auxiliary load-balancing or $z$-loss constraints.

\textbf{Depth-Wise Modality Convergence.} Expert specialization varies strongly across model depth (Fig.~\ref{fig:expert-routing}). Early decoder layers exhibit high modality divergence, with most active experts dedicated primarily to either vision or text. In deeper layers, this separation steadily fades as representations across all modalities converge toward a compact set of shared experts, before briefly re-diverging at the final layer to form the output embedding.

\section{Additional Details for Adaptive Computation}
\label{sec:adaptive-details}

This section gives implementation details for every adaptive-computation method
of \S\ref{sec:training-efficiency} and Tab.~\ref{tab:inference}, grouped by
whether the decision is learned during training or taken at inference.

\subsection{Training-Based Methods}
\label{sec:trained-arms}
\label{sec:adaptive-training-setup}

All training-based variants extend the contrastive framework described in Appendix~\ref{sec:implementation-details}. Models are trained on 64 NVIDIA H200 GPUs (expert parallel size 64) in BF16 precision using AdamW ($\text{weight decay} = 0.01$, gradient clip norm $1.0$, logit scale $50$, maximum sequence length $8\text{,}192$). LoRA adapters ($r=64, \alpha=128$, dropout $0.05$) are applied to visual and language projection layers (\texttt{linear\_qkv}, \texttt{linear\_proj}, \texttt{linear\_fc1}, \texttt{linear\_fc2}). Global batch size is set to $512$, except for GRPO, which uses a global batch size of $256$ with $8$ rollouts per input.

\textbf{Initialization and Distillation Regimes.}
Token pruning, layer skipping, AdaMoE, and GRPO are initialized from the pretrained \method{}-A3B checkpoint and fine-tuned for one epoch. Token pruning and layer skipping train with a learning rate of $10^{-5}$ and an $L_2$ distillation loss ($\text{weight} = 1.0$) against cached full-compute teacher hidden states to stabilize representations during gate optimization. AdaMoE is fine-tuned at a learning rate of $10^{-4}$ without distillation. GRPO uses a learning rate of $5 \times 10^{-5}$ and integrates distillation directly into its reward function. Expert skipping is initialized directly from the base backbone and trained jointly with contrastive adaptation over two epochs at a learning rate of $10^{-4}$ without distillation.

\textbf{Token Pruning.}
Pruning decisions are evaluated after decoder block~2, targeting a $30\%$ visual token retention rate via an $L_1$ budget loss ($\text{weight} = 0.5$, $15\%$ warmup). The Gumbel-Softmax temperature is linearly annealed from $4.0$ to $0.5$ over $2\text{,}000$ steps, with predictor weights updated at $40\times$ the backbone learning rate. Pruning is restricted exclusively to visual tokens.

\textbf{Expert Skipping.}
Expert skipping penalizes active compute via an $L_1$ penalty weight of $0.02$ on $(1 - \rho_{\text{skip}})$. This penalty is held at zero for the first $15\%$ of training steps and linearly ramped to full weight over the subsequent $5\%$ of steps.

\textbf{Layer Skipping.}
A gate module at each block evaluates the hidden state at the \texttt{<emb>} position to produce a binary keep-or-skip decision for the entire block. When a block is skipped, its residual passes through unmodified, bypassing both attention and MoE sub-layers. The gate is trained via a straight-through estimator targeting a $20\%$ skip rate under unit loss weight, applied from decoder block~3 onward. In practice, the policy converges to a $20.5\%$ empirical skip rate concentrated in 9 decoder blocks, with the remaining eligible blocks left unskipped.

\textbf{Adaptive Top-$k$.}
Following AdaMoE \citep{zeng2024adamoe}, we augment the router with 256 null experts alongside the 256 physical experts and perform a joint top-8 selection across all 512 candidates. A token's effective width corresponds to the number of top-8 slots allocated to physical experts, eliminating manual threshold tuning. A load-balancing penalty ($\text{weight} = 0.02$) maintains the mean activation near four physical experts during the first half of training, after which it drops to $10^{-4}$ to allow unconstrained routing convergence. The policy settles at an average of 2.94 physical experts per token at evaluation ($63.3\%$ reduction in routed expert execution); we compare this setting against MoDES at an equivalent target budget in \S\ref{sec:inference-results}.

\textbf{Group Relative Policy Optimization (GRPO).}
\label{sec:controller-details}
Differentiable training (\textit{e.g.} STE) is unstable (App.~\ref{sec:failed-adaptivity}).  We
instead treat each gate's decision as a policy and adapt GRPO~\citep{shao2024deepseekmath},
which samples the discrete action in place of differentiating through it. Each efficiency gate (\textit{e.g.} an expert skipping gate) predicts a distribution over its actions. For each input, we sample a group of outputs $\{a_i\}_{i=1}^{G}$, where each rollout samples one action from every gate, producing an embedding $e_{a_i}$. As adaptive computation has no pretrained reference policy to regularize toward, we replace the KL term with a distillation loss toward a teacher embedding $e^{\mathrm{full}}$, computed by a full-compute greedy forward pass.  To avoid reward hacking, we gate the efficiency reward $S(a_i)$ on the success of the in-batch retrieval $\mathrm{recall}@G(a_i)\in[0,1]$:
\begin{equation*}
\begin{gathered}
\mathcal{L} = -\,\mathbb{E}\!\left[\frac{1}{G}\sum_{i=1}^{G}
\min\!\Big(r_i\hat A_i,\,\mathrm{clip}\big(r_i,1{-}\epsilon,1{+}\epsilon\big)\hat A_i\Big)\right]
+ \lambda_{\mathrm{d}}\,\mathbb{E}\big[1-\cos(e_a,e^{\mathrm{full}})\big],
\\[4pt]
r_i=\frac{\pi_\theta(a_i)}{\pi_{\theta_{\mathrm{old}}}(a_i)},
\;\;
\hat A_i=\frac{R(a_i)-R_{\mu}}{R_{\sigma}},
\;\;
R(a_i) = \underbrace{\mathrm{recall}@G(a_i)}_{\text{recall gate}}\big(
\underbrace{\bar{s}^{+}(a_i) - \bar{s}^{-}(a_i)}_{\text{ranking margin}}
+ \lambda_{\mathrm{s}} S(a_i)\big),
\end{gathered}
\end{equation*}
where $R_{\mu},R_{\sigma}$ are the mean and standard deviation of the reward over the group. $\bar{s}^{+}(a_i)$ and $\bar{s}^{-}(a_i)$ denote the similarity between positive and negative pairs.

GRPO optimizes a lightweight three-way classification head per MoE layer that emits categorical decisions over $\{\textsc{route}, \textsc{skip}, \textsc{drop}\}$ for each token. Native expert routing remains greedy, meaning the policy dictates only execution state without altering expert selection logic. Auxiliary routing losses are disabled.

The policy is trained on 8 rollouts per sequence sampled at temperature $1.0$ using a learning rate of $5 \times 10^{-5}$ without KL penalties or value baselines, utilizing group-standardized advantages across rollouts. Log-probabilities sum categorical policy decisions across active tokens and MoE layers, normalized by the total decision count to equalize variable sequence lengths. To manage memory overhead, gradients are replayed one rollout at a time. Updates are computed on-policy ($r^{(i)} \equiv 1$, rendering ratio clipping inactive). To avoid distorting core feature spaces, policy gradients flow exclusively to the efficiency gate heads; base backbone and LoRA weights remain fixed at their contrastive values \citep{lan2025umer1,jiang2026embedrl}.

The overall reward combines an efficiency score $S(a)$ and a cosine distillation reward against teacher embeddings ($\text{weight} = 1.0$, softmax temperature $0.2$ for negative pairs). The efficiency reward credits each action by the compute it removes at each layer:
\begin{equation}
S(a) = 1 - \frac{\mathbb{E}[\mathrm{FLOPs}(a)]}{\mathrm{FLOPs}_{\mathrm{dense}}},
\end{equation}
\begin{equation}
\mathbb{E}[\mathrm{FLOPs}(a)] = \sum_{\ell} \sum_{t} \alpha_{t,\ell} \left[ f^{\mathrm{attn}}_\ell + f_{\mathrm{sh}} + (1 - p^{\mathrm{skip}}_{t,\ell} - p^{\mathrm{drop}}_{t,\ell}) f_{\mathrm{rt}} \right],
\end{equation}
where $\alpha_{t,\ell}$ is the probability that token $t$ reaches layer $\ell$, and $f^{\mathrm{attn}}_\ell, f_{\mathrm{sh}}, f_{\mathrm{rt}}$ represent layer-specific attention, shared expert, and routed expert FLOP costs. Expert skipping removes $f_{\mathrm{rt}}$ at that layer and token dropping removes all downstream per-token compute.

\textbf{Explored Variants and Negative Results.}
\label{sec:failed-adaptivity}
We additionally evaluated A-ViT and PonderNet-style dynamic halting mechanisms \citep{yin2022avit,banino2021pondernet} based on learned hidden states and GDN-$\beta$ token scores. Training combined a ponder penalty and a geometric distribution prior over steps under a joint Gumbel-Softmax / STE loss formulation \citep{jang2017gumbel}:
\begin{equation*}
\mathcal{L}_{\mathrm{halt}} = \mathcal{L}_{\mathrm{emb}} + \lambda_{\mathrm{p}} \mathbb{E}[N_{\mathrm{halt}}] + \lambda_{\mathrm{kl}} \mathrm{KL}\left(q_{\mathrm{halt}} \,\|\, \mathrm{Geom}(\gamma)\right).
\end{equation*}
Table~\ref{tab:failed-adaptivity} summarizes the observed failure modes. Aggressive regularization degraded target representation quality, whereas conservative settings caused gate gradients to saturate near zero reduction. Joint pruning techniques such as DiEP \citep{bai2025diep} similarly failed to achieve competitive accuracy-efficiency trade-offs. We hypothesize that credit assignment through non-differentiable or delayed attention paths introduces high gradient variance during early training, causing efficiency penalties to dominate before representation alignment can stabilize.

\begin{table}[ht]
\centering
\caption{Summary of non-convergent or suboptimal baseline implementations.}
\label{tab:failed-adaptivity}
\small
\begin{tabularx}{\textwidth}{lX}
\toprule
Variant & Observed behavior \\
\midrule
Projected A-ViT gate & Training is initially healthy, then drop exceeds roughly
85\%; the embedding token loses visual context and both representation and gradients
collapse. \\
Cumulative drop logit & The compute reward drives nearly all halting to the
first layer; the distribution prior does not offset the incentive to drop
earliest. \\
GDN-$\beta$ halting & With a stable negative sign (low $\beta$ drops first), the
gate frequently becomes inert and returns toward zero drop; more aggressive
settings destabilize training. \\
Learned exit-halting head & Exit probabilities converge to the final layer, producing
negligible savings under stable quality. \\
\bottomrule
\end{tabularx}
\end{table}

\subsection{Inference-Based Methods}
\label{sec:inference-details}

This subsection details the training-free dynamic efficiency heuristics evaluated in \S\ref{sec:inference-eff}.

\textbf{Token Pruning via GDN-$\beta$.}
Visual tokens are ranked using the Gated-DeltaNet write strength $\beta$ evaluated at decoder block~4. The lowest-scoring tokens are pruned, and remaining sequences are gathered to reduce compute across all subsequent blocks. Text tokens are fully preserved.

\textbf{Expert Skipping via GDN-$\beta$.}
Using the same $\beta$ signal, tokens ranking in the lower fraction per MoE block skip routed experts.

\textbf{Static Expert Pruning.} Methods including REAP~\citep{lasby2025reap}, MAN~\citep{liu2026scoreexperts}, MSAN~\citep{liu2026scoreexperts}, and router logit thresholding evaluate expert importance scores over a static calibration set of 16 balanced samples. The lowest-ranked experts are permanently removed across all layers. Because routing maintains top-8 selection over reduced expert pools, this strategy reduces total parameters without reducing dynamic FLOPs per token.

\textbf{Adaptive Top-$k$ Routing.}
\label{sec:topk-probe}
Uniform top-$k$ keeps the $k$ highest-scoring of the native eight experts at
every layer, with no calibration and no learned gate.  The pretrained router's
ranking is what makes the reduced budget cheap: at $k=4$ accuracy falls by
$0.5$ points, while drawing four of the eight at random costs $6.6$ at the
identical $11.6\%$ FLOPs reduction (Tab.~\ref{tab:inference}).  The ordering
the router induces over its top-8 therefore accounts for most of what a halved
budget preserves.

\textbf{MoDES Calibration.}
\label{sec:modes-calibration}
MoDES derives layer-wise scaling parameters $s_\ell$ and a global activation threshold $\tau$ using a calibration subset from the ImageNet-1K evaluation split.

\textbf{Adaptive Otsu Thresholding for Expert Skipping.}
Rather than enforcing uniform skip ratios across layers, we explore dynamically computing per-layer thresholds using Otsu's method \citep{otsu1979threshold}. For each layer, the score distribution is partitioned into two classes by maximizing inter-class variance; tokens in the lower partition skip routed experts. We evaluate two scoring signals: (1) GDN write strength $\beta$ across 30 GDN layers, and (2) maximum pre-softmax router logits across all 40 MoE layers. A dispersion threshold (coefficient of variation $> 0.05$--$0.40$) restricts partitioning to layers with sufficiently bimodal score distributions, subject to a maximum $70\%$ per-layer skip cap. Neither variant requires pre-calibration.

\newpage
\section{Additional Analysis on GDN Write Strength $\beta$}
\label{sec:additional-efficiency}

\label{sec:beta-depth}
\label{sec:gdn-beta-details}

\begin{wrapfigure}{r}{0.52\textwidth}
\centering
\vspace{-10pt}
  \includegraphics[width=\linewidth]{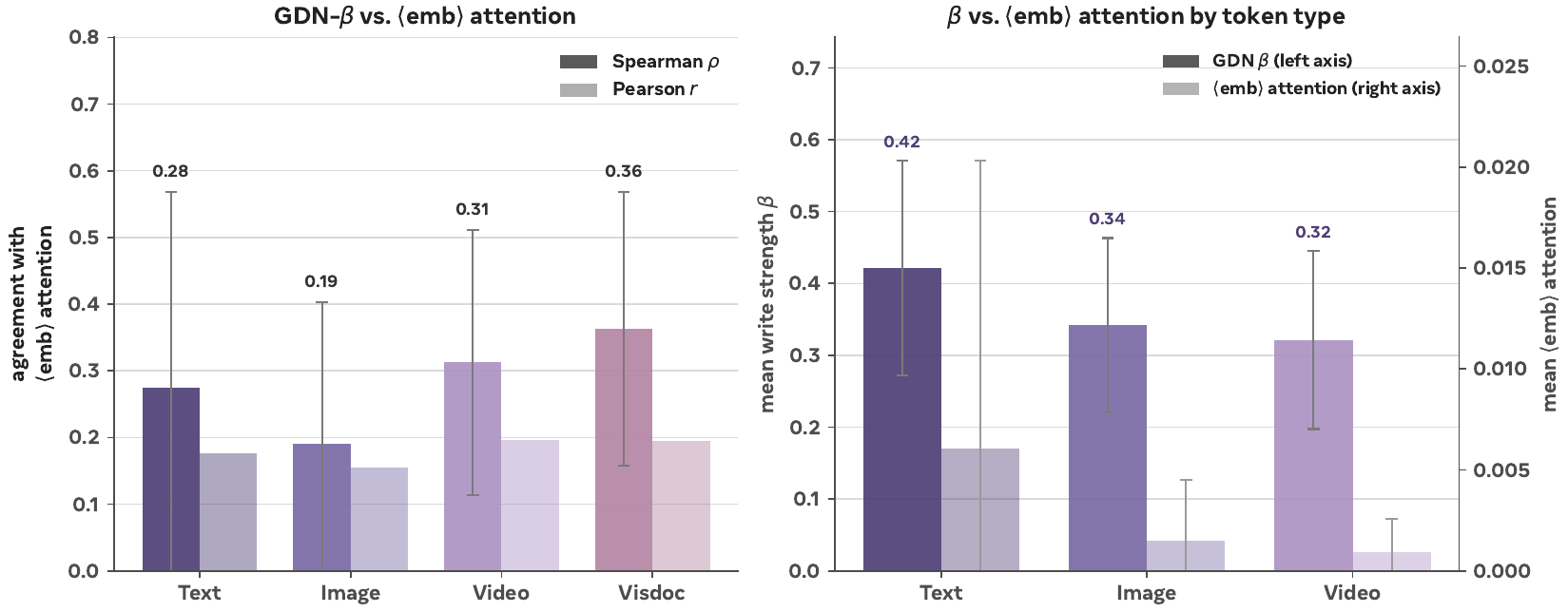}

\captionsetup{font=footnotesize}
  \caption{$\beta$ \textit{vs.} attention score.  Left: per-modality agreement (Spearman $\rho$ and
Pearson $r$); Right: mean $\beta$
(left axis) and mean \texttt{<emb>} attention (right axis) by token type.}
\label{fig:gdn-beta}
\vspace{-5pt}
\end{wrapfigure}

\textbf{$\beta$ functions as an independent salience signal.}
GDN $\beta$ provides an informative token pruning signal that operates independently of standard attention mechanisms. While attention distributions are spiky and heavily concentrated, $\beta$ tends to be more spatially and temporally smooth. Moreover, we find $\beta$ to have a more even focus across different modalities in \method{}, compared to attention scores, which are mostly concentrated on text tokens. Despite $\beta$'s  distinct properties from attention scores, pruning tokens by $\beta$ performs comparably to the trained token pruning gate (Fig.~\ref{fig:budget-accuracy}), confirming its utility as an effective salience metric.  The remainder of this section reports further analysis of $\beta$: its per-layer profile, its agreement with the \texttt{<emb>} attention row and with received attention mass, and its spread by token type.

\textbf{Depth Dynamics and Attention Comparison.}
We evaluate the layer-wise behavior of GDN write strength $\beta$ using three representative samples per modality. Across all evaluated tokens, the layer-wise mean of $\beta$ spans $[0.196, 0.596]$, with coefficients of variation ranging from $0.065$ to $0.496$. In an eight-example diagnostic pairing each of the 10 GDN layers immediately preceding a full-attention layer with its corresponding attention block, the Spearman correlation between $\beta$ and incoming attention mass reaches $0.546$, $0.619$, $0.491$, and $0.524$ for image, text, video, and VisDoc modalities, respectively. However, attention allocation exhibits considerably higher spatial variance, yielding a mean token-wise coefficient of variation of $2.316$, compared to $0.226$ for $\beta$. 

\begin{figure}[htbp]\centering
  \includegraphics[width=\textwidth]{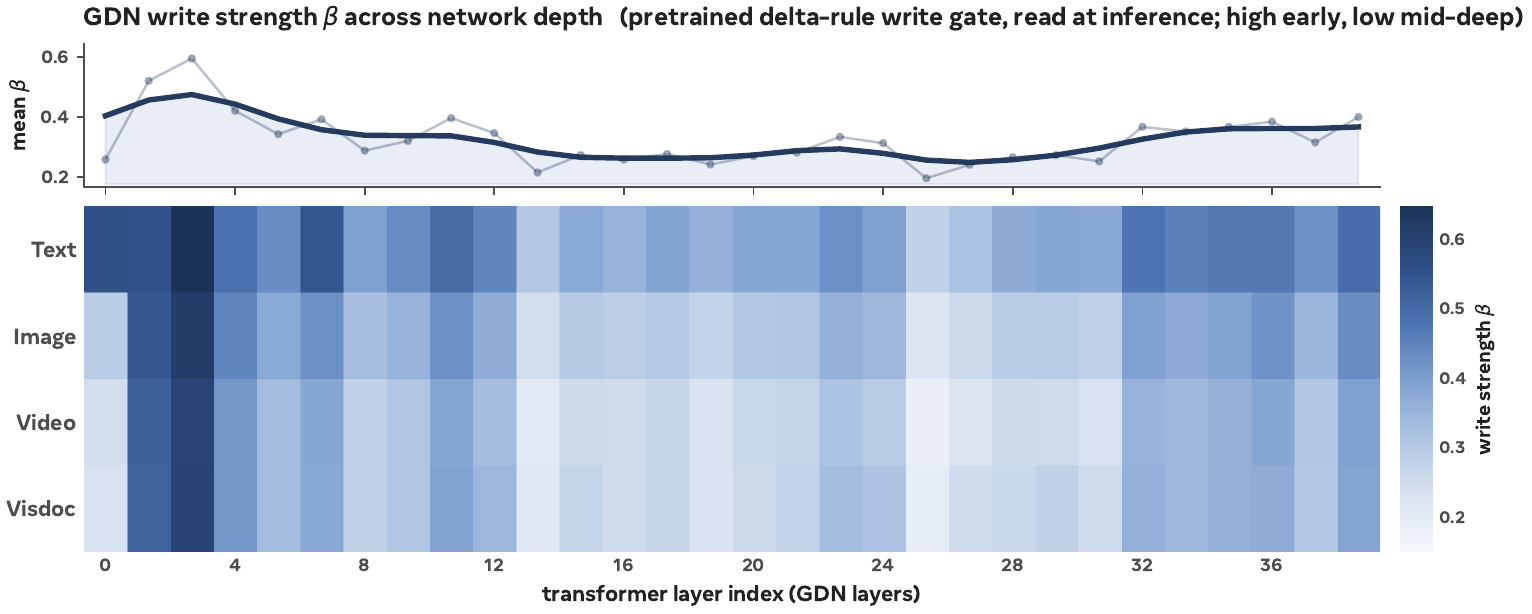}
  \caption{\textbf{The GDN write strength is depth-structured.}  Per-token
  Gated-DeltaNet delta-rule write strength $\beta$ (a gate of the pretrained
  backbone, read at inference), averaged per layer (top: network-wide marginal;
  bottom: per modality).  Tokens write strongly in the early blocks and weakly
  through the middle-to-deep blocks.  The middle panel of
  Fig.~\ref{fig:depth-allocation} plots the same quantity beside the two learned
  actions.}
  \label{fig:skip-depth}
\end{figure}
\begin{figure}[htbp]\centering
  \includegraphics[width=\textwidth]{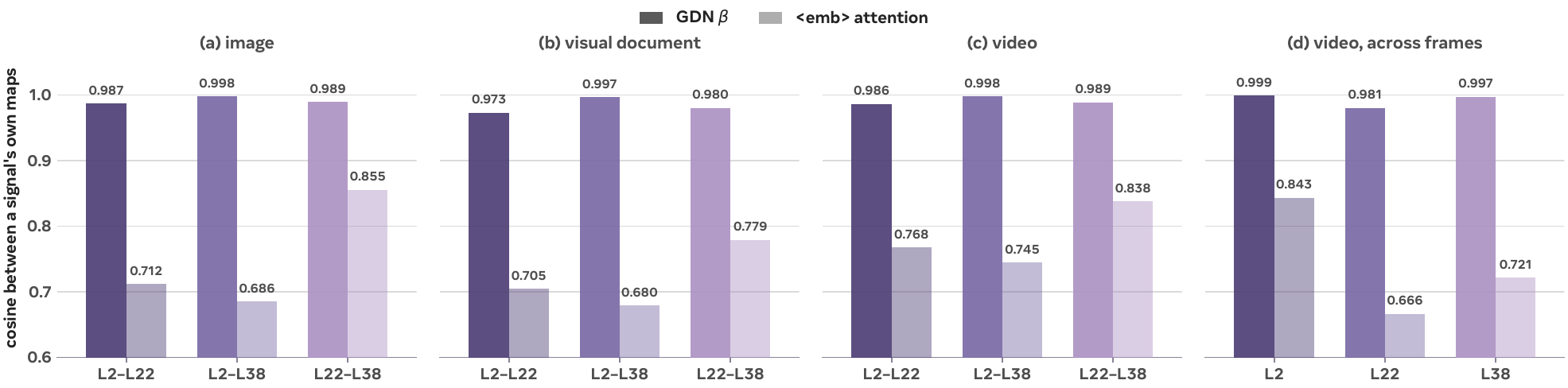}
  \caption{Figures (a)--(c): cosine similarity across layers; Figure (d) compares cosine similarity across video frames within one layer.}
  \label{fig:beta-consistency}
\end{figure}
\begin{figure}[htbp]\centering
  \includegraphics[width=\textwidth]{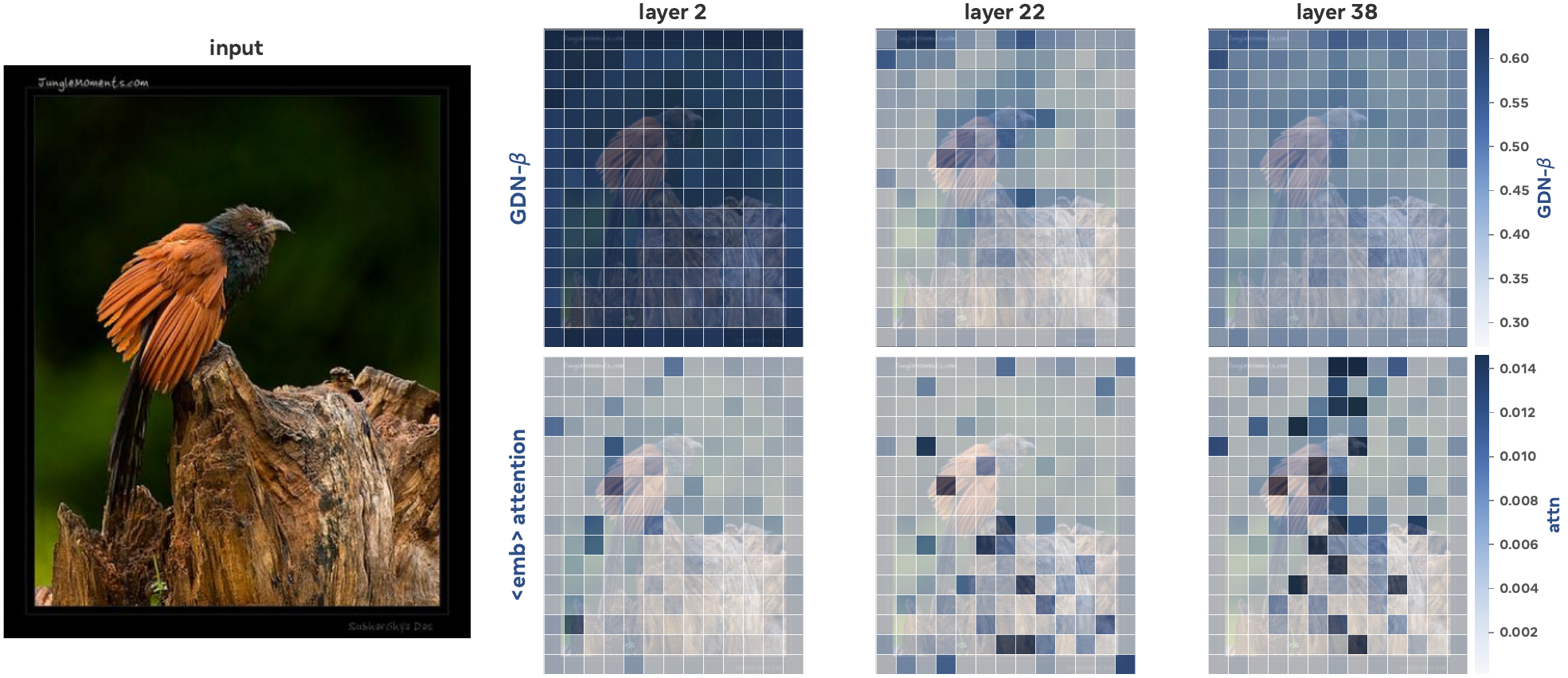}
  \caption{\textbf{Image: GDN-$\beta$ vs.\ the \texttt{<emb>} attention.}
  The clean input is at left; the panels overlay GDN-$\beta$ (top) and the
  \texttt{<emb>} token's attention onto the visual tokens (bottom) at layers 2,
  22, and 38.  Cell colour encodes magnitude against the colourbars, with the
  image kept faintly underneath for spatial reference.  $\beta$ is strong and
  near-uniform early and localizes smoothly with depth, whereas the \texttt{<emb>}
  attention is sparse and scattered and only strengthens in the deeper layers.}
  \label{fig:qual-image-emb}
\end{figure}
\begin{figure}[htbp]\centering
  \includegraphics[width=\textwidth]{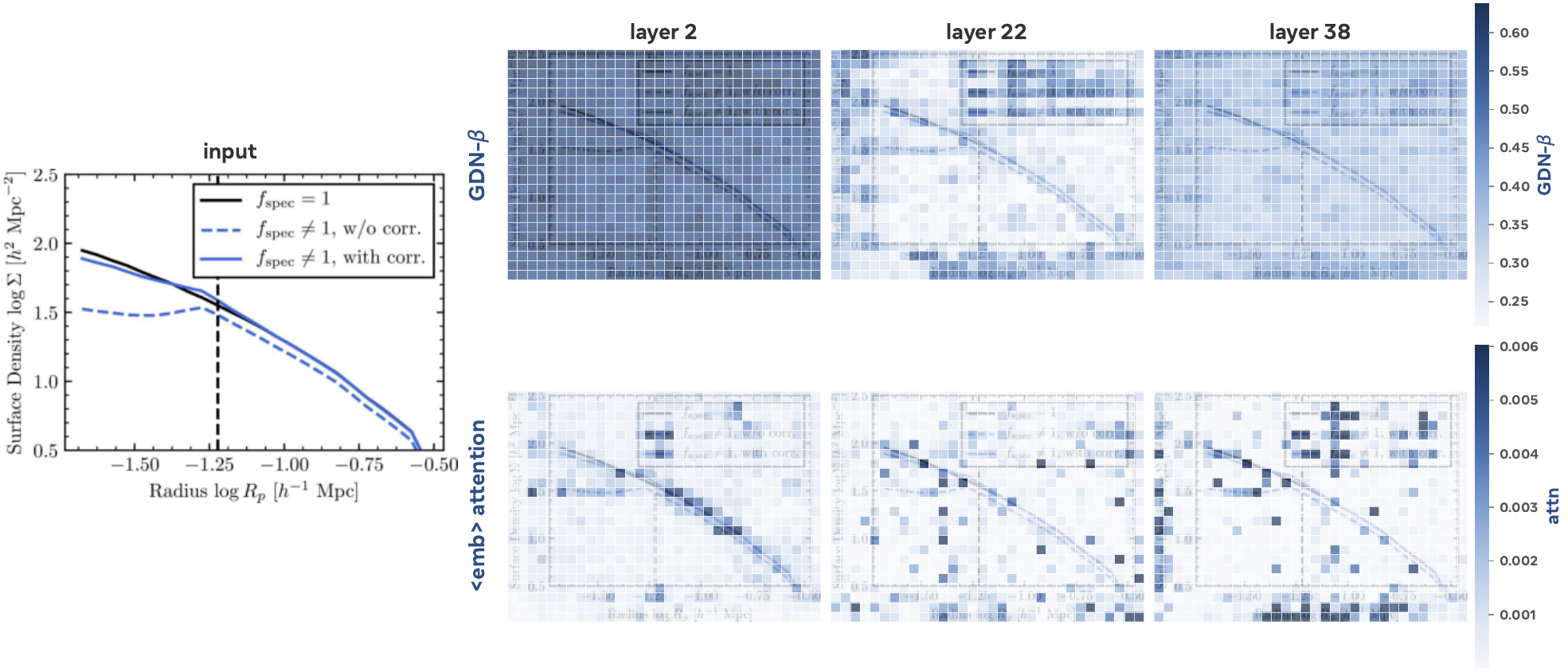}
  \caption{\textbf{Visual document: GDN-$\beta$ vs.\ the \texttt{<emb>} attention.}  Same layout as Fig.~\ref{fig:qual-image-emb}, on a document
  page.  As on natural images, $\beta$ is the smoother, more distributed signal
  across depth.}
  \label{fig:qual-visdoc-emb}
\end{figure}
\begin{figure}[htbp]\centering
  \includegraphics[width=\textwidth]{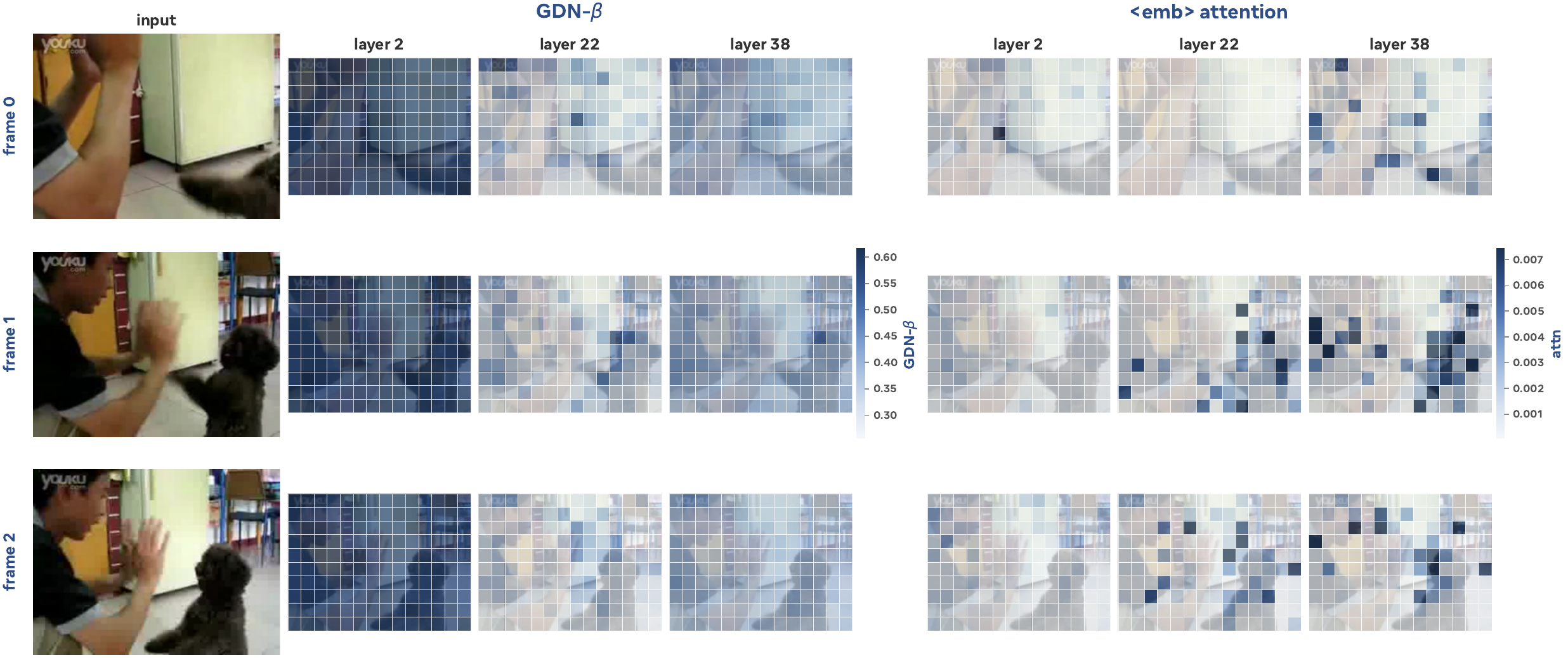}
  \caption{\textbf{Video: GDN-$\beta$ vs.\ the \texttt{<emb>} attention.}
  Rows are frames: the clean frame at left, then GDN-$\beta$ and the
  \texttt{<emb>} attention side by side, each at layers 2, 22, and 38.
  The \texttt{<emb>} attention is nearly empty at layer~2 and
  sharpens into scattered per-frame peaks with depth, whereas $\beta$ is strong
  early and spatially smooth, so the two signals carry complementary information.}
  \label{fig:qual-video-emb}
\end{figure}

\textbf{Spatial and Temporal Consistency.}
Qualitative overlays (Figures~\ref{fig:qual-image-emb}--\ref{fig:qual-video-emb}) and layer-wise comparisons across blocks 2, 22, and 38 (Figure~\ref{fig:beta-consistency}) demonstrate three key properties of $\beta$:
\begin{enumerate}
    \item Semantic Alignment: $\beta$ highlights semantically informative regions—such as foreground objects in natural images or plot/text elements in documents—forming contiguous spatial clusters rather than isolated activation noise.
    \item Depth Stability: $\beta$ distributions remain stable across network layers, whereas attention maps shift dynamically. Computing cross-layer cosine similarity yields an average of $0.989$ (minimum $0.973$) for $\beta$, compared to $0.752$ (range $[0.68, 0.86]$) for \texttt{<emb>} attention. Tokens identified as high-value in early blocks preserve high scores deeper in the network, even as attention redirects focus.
    \item Temporal Persistence: Across video frames within a given layer, $\beta$ maps exhibit high cross-frame cosine similarity ($0.992$ vs.\ $0.744$ for attention).
\end{enumerate}
This spatial and temporal stability makes $\beta$ a reliable candidate for early single-shot pruning, ensuring that token removal decisions remain valid throughout subsequent layers.

\begin{wraptable}{r}{0.47\textwidth}
\vspace{-1.1\baselineskip}
\centering
\caption{Inference-only GDN-$\beta$ decision-depth diagnostic.  Layer indices
are zero-based.}
\label{tab:beta-depth}
\small
\setlength{\tabcolsep}{4.5pt}
\begin{tabular}{lrrrr}
\toprule
Keep & layer & Accuracy & $\Delta$ & FLOPs $\downarrow$ \\
\midrule
75\% & 2 & 70.07 & $-0.25$ & 6.47\% \\
75\% & 4 & 70.03 & $-0.29$ & 6.12\% \\
25\% & 2 & 66.26 & $-4.06$ & 19.36\% \\
25\% & 4 & 66.05 & $-4.27$ & 18.33\% \\
\bottomrule
\end{tabular}
\vspace{-1.0\baselineskip}
\end{wraptable}

\textbf{Pruning Decision Depth.}
Table~\ref{tab:beta-depth} evaluates sensitivity to the layer at which the single-shot pruning mask is generated. Executing the decision at layer~2 rather than layer~4 alters accuracy by $+0.04$ at a $75\%$ keep rate and $+0.21$ at a $25\%$ keep rate, yielding modest incremental FLOP savings of $0.35$ and $1.03$ percentage points, respectively. These minor variations indicate that pruning effectiveness is broadly insensitive to the exact decision layer.

\begin{figure}[htbp]\centering
  \includegraphics[width=0.86\textwidth]{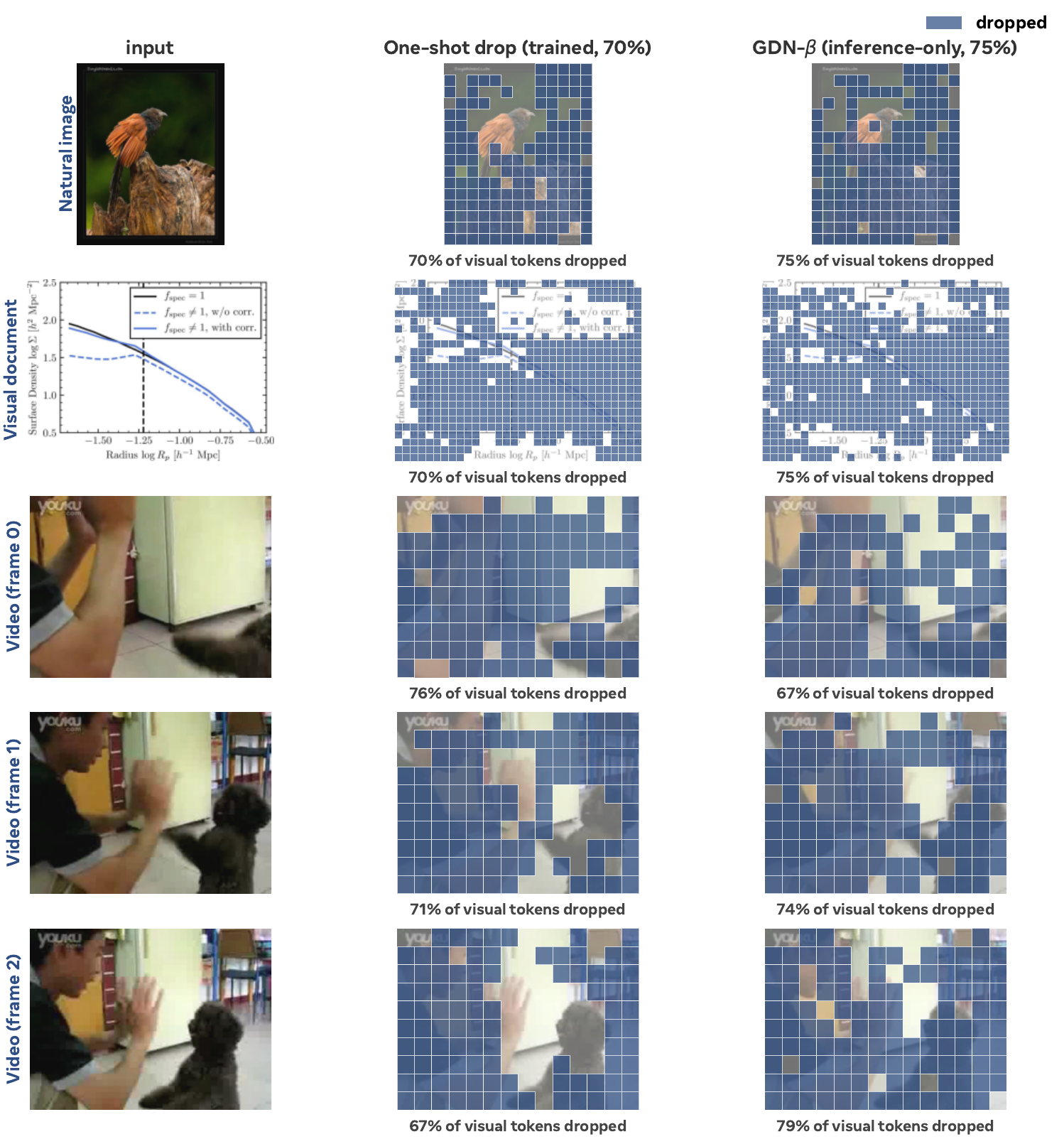}
  \caption{Trained vs.\ inference-only token dropping, at an equal budget.}
  \label{fig:trained-drop}
\end{figure}

\begin{figure}[htbp]\centering
  \includegraphics[width=0.86\textwidth]{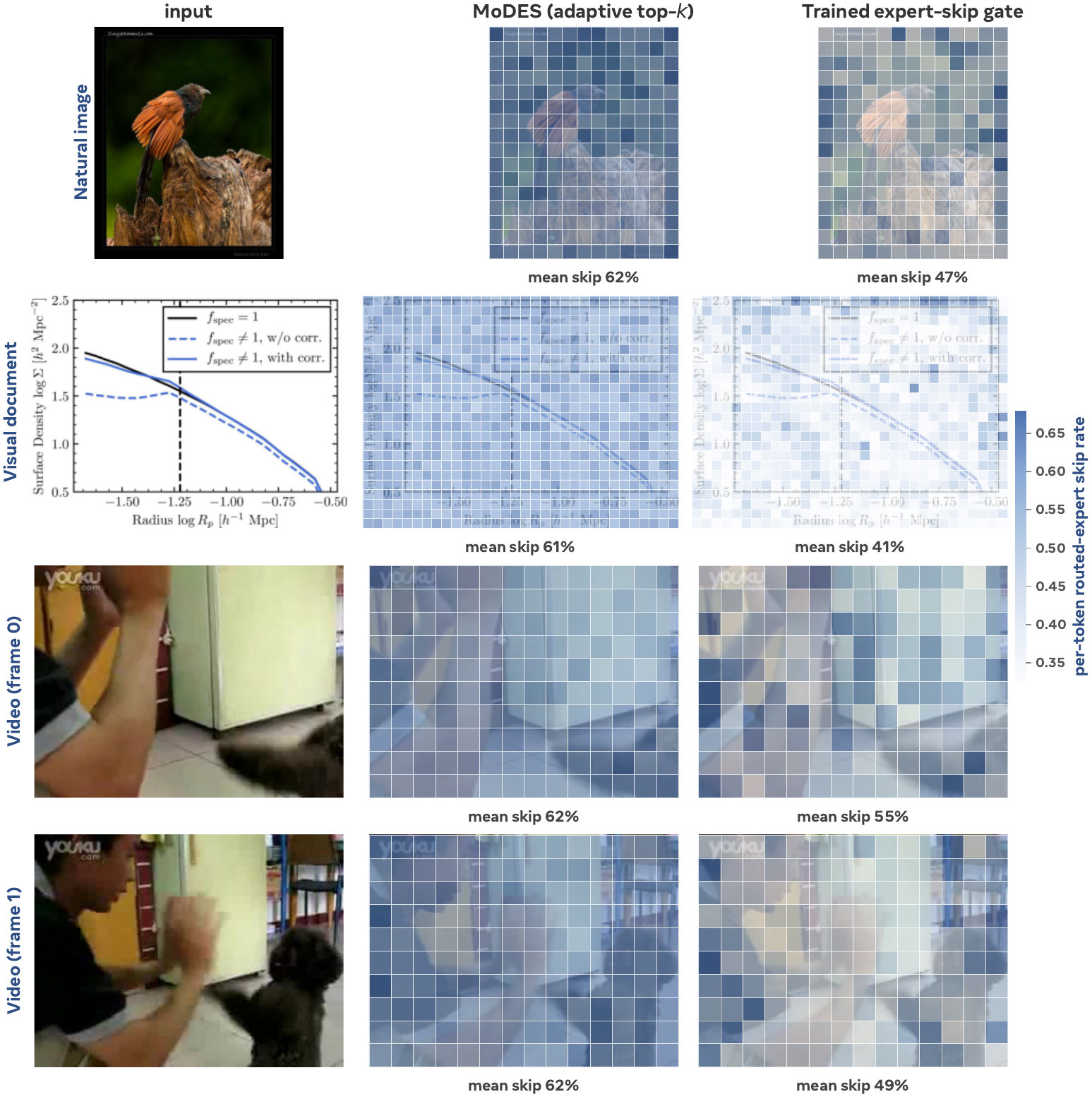}
  \caption{\textbf{How the two expert-skipping policies allocate their budget.}
  Per-token fraction of the 40 MoE layers whose routed-expert branch is skipped,
  for MoDES (adaptive top-$k$, inference-only) and the trained expert skipping, at the same overall skip rate.}
  \label{fig:trained-skip}
\end{figure}

\textbf{Visual Comparison of Trained and Inference-Only Token Pruning and Expert skipping Methods.}
Figures~\ref{fig:trained-drop} and~\ref{fig:trained-skip} compare trained adaptive computing methods (token pruning and adaptive top-$k$ / expert skipping) with their inference-only counterparts evaluated on identical inputs. As shown in Fig.~\ref{fig:trained-drop}, the single-shot token dropper and the GDN-$\beta$ policy select nearly identical token subsets. Conversely, Fig.~\ref{fig:trained-skip} reveals that the trained expert-skipping gate and adaptive top-$k$ routing diverge notably in their allocation decisions.

\FloatBarrier

\clearpage
\newpage

\section{Per-Task MMEB-V2 Results}
\label{sec:per-task}

{\scriptsize
\setlength{\tabcolsep}{1.6pt}\renewcommand{\arraystretch}{1.02}
\begin{longtable}{@{}l rrrrrrrr @{\hspace{4pt}} rr@{}}
\caption{Per-task MMEB-V2 scores.}\label{tab:per-task-backbones}\\
\toprule
& \multicolumn{3}{c}{\textbf{Single-pass}} & \multicolumn{5}{c}{\textbf{TTE-based}} & \multicolumn{2}{c}{\textbf{\method{}}}\\
\cmidrule(lr){2-4}\cmidrule(lr){5-9}\cmidrule(lr){10-11}
\textbf{Task} & \textbf{\shortstack[c]{VLM2Vec\\V2-2B}} & \textbf{\shortstack[c]{UniME\\V2-7B}} & \textbf{\shortstack[c]{\\BToks}} & \textbf{\shortstack[c]{\\PLUME}} & \textbf{\shortstack[c]{UME-R1\\2B}} & \textbf{\shortstack[c]{Embed\\RL-2B}} & \textbf{\shortstack[c]{UME-R1\\7B}} & \textbf{\shortstack[c]{Embed\\RL-4B}} & \textbf{\shortstack[c]{\\A3B}} & \textbf{\shortstack[c]{\\A10B}} \\
\midrule
\endfirsthead
\multicolumn{11}{c}{\tablename\ \thetable{} -- continued from previous page}\\
\toprule
& \multicolumn{3}{c}{\textbf{Single-pass}} & \multicolumn{5}{c}{\textbf{TTE-based}} & \multicolumn{2}{c}{\textbf{\method{}}}\\
\cmidrule(lr){2-4}\cmidrule(lr){5-9}\cmidrule(lr){10-11}
\textbf{Task} & \textbf{\shortstack[c]{VLM2Vec\\V2-2B}} & \textbf{\shortstack[c]{UniME\\V2-7B}} & \textbf{\shortstack[c]{\\BToks}} & \textbf{\shortstack[c]{\\PLUME}} & \textbf{\shortstack[c]{UME-R1\\2B}} & \textbf{\shortstack[c]{Embed\\RL-2B}} & \textbf{\shortstack[c]{UME-R1\\7B}} & \textbf{\shortstack[c]{Embed\\RL-4B}} & \textbf{\shortstack[c]{\\A3B}} & \textbf{\shortstack[c]{\\A10B}} \\
\midrule
\endhead
\midrule \multicolumn{11}{r}{\scriptsize Continued on next page}\\
\endfoot
\bottomrule
\endlastfoot
\addlinespace[2pt]
\rowcolor{plmgray}\textbf{Image classification}\ \ \scriptsize(10 tasks) & 62.9 & 65.6 & 64.3 & 66.5 & 64.8 & 62.8 & \textbf{67.1} & 63.7 & 60.6 & 64.4 \\[1pt]
Country211 & 25.8 & 19.0 & 25.2 & 25.0 & 23.4 & 20.0 & 25.0 & 19.4 & 21.4 & \textbf{26.3} \\
HatefulMemes & 55.8 & 66.1 & 62.5 & 75.5 & 75.2 & 65.0 & \textbf{79.0} & 66.2 & 59.8 & 69.8 \\
ImageNet-1K & \textbf{80.8} & 78.9 & 80.5 & 74.1 & 75.3 & 78.0 & 80.4 & 79.5 & 78.1 & 79.5 \\
ImageNet-A & 47.6 & 49.0 & 46.9 & 50.8 & 50.4 & \textbf{59.2} & 53.9 & 58.1 & 52.9 & 55.2 \\
ImageNet-R & 89.3 & 89.1 & 85.6 & 87.5 & 88.7 & 88.5 & \textbf{90.1} & 88.2 & 87.9 & 88.4 \\
N24News & 73.0 & 66.0 & 74.0 & 81.1 & 81.1 & 44.9 & \textbf{82.3} & 48.3 & 39.5 & 50.5 \\
ObjectNet & 65.1 & 73.2 & 68.5 & 61.5 & 52.0 & 74.8 & 42.3 & \textbf{75.4} & 69.1 & 73.7 \\
Place365 & 36.1 & 43.8 & 38.7 & 42.4 & 42.6 & 43.9 & \textbf{46.8} & 43.1 & 44.1 & 45.3 \\
SUN397 & 70.9 & 78.8 & 74.9 & 76.9 & 79.4 & 75.4 & \textbf{80.3} & 79.2 & 72.9 & 74.3 \\
VOC2007 & 84.9 & \textbf{92.5} & 85.7 & 86.1 & 80.0 & 78.7 & 90.8 & 79.5 & 80.8 & 81.3 \\
\addlinespace[2pt]
\rowcolor{plmgray}\textbf{Image VQA}\ \ \scriptsize(10 tasks) & 56.4 & 68.7 & 59.8 & 59.2 & 62.8 & 67.9 & 69.2 & 70.5 & 73.8 & \textbf{75.4} \\[1pt]
A-OKVQA & 44.0 & \textbf{71.5} & 48.6 & 49.9 & 51.1 & 54.7 & 58.7 & 59.3 & 63.1 & 64.6 \\
ChartQA & 48.1 & 59.2 & 51.2 & 49.8 & 64.9 & 80.7 & 75.1 & 80.9 & 79.0 & \textbf{82.5} \\
DocVQA & 90.1 & 92.4 & 91.9 & 89.9 & 92.2 & 92.4 & 93.8 & 94.3 & \textbf{96.0} & \textbf{96.0} \\
GQA & 65.4 & 69.0 & 63.9 & 69.1 & 67.3 & 64.9 & 69.3 & 68.5 & 76.3 & \textbf{77.6} \\
InfographicsVQA & 59.1 & 67.1 & 61.6 & 59.6 & 67.7 & 76.7 & 79.2 & 77.5 & 80.0 & \textbf{82.6} \\
OK-VQA & 51.7 & 71.7 & 61.8 & 60.5 & 62.4 & 61.4 & 71.7 & 67.3 & 73.9 & \textbf{75.1} \\
ScienceQA & 38.1 & 55.2 & 40.1 & 42.9 & 42.7 & 57.3 & 53.7 & 61.6 & 63.6 & \textbf{67.4} \\
TextVQA & 71.6 & 84.4 & 79.5 & 78.9 & 78.6 & 83.8 & 83.5 & 84.3 & 88.5 & \textbf{89.6} \\
Visual7W & 52.8 & \textbf{62.7} & 49.2 & 47.6 & 54.1 & 52.7 & 55.2 & 55.3 & 59.7 & 58.8 \\
VizWiz & 43.3 & 53.4 & 49.8 & 46.5 & 46.8 & 54.5 & 51.6 & 56.2 & 58.0 & \textbf{60.1} \\
\addlinespace[2pt]
\rowcolor{plmgray}\textbf{Text-to-image retrieval}\ \ \scriptsize(6 tasks) & 79.3 & 82.0 & 78.6 & 76.2 & 77.2 & 79.2 & 81.9 & 80.8 & 83.2 & \textbf{85.4} \\[1pt]
EDIS & 84.2 & 88.5 & 85.4 & 81.8 & 88.0 & 84.5 & 92.0 & 87.4 & 91.9 & \textbf{92.9} \\
MSCOCO\_t2i & 75.9 & \textbf{80.0} & 73.9 & 74.1 & 75.1 & 79.4 & 78.3 & 78.9 & 78.7 & 79.7 \\
VisDial & 82.7 & 84.8 & 78.4 & 72.6 & 76.6 & 81.5 & 80.7 & 84.9 & 85.1 & \textbf{88.4} \\
VisualNews\_t2i & 74.7 & 77.3 & 72.7 & 71.3 & 71.7 & 71.9 & 76.8 & 73.7 & 76.4 & \textbf{81.9} \\
WebQA & 90.6 & 90.2 & 90.5 & 89.1 & 90.0 & 89.3 & \textbf{90.9} & 90.5 & 89.1 & \textbf{90.9} \\
Wiki-SS-NQ & 67.6 & 70.9 & 70.5 & 68.6 & 62.0 & 68.9 & 72.5 & 69.6 & 77.8 & \textbf{78.8} \\
\addlinespace[2pt]
\rowcolor{plmgray}\textbf{Image-to-text retrieval}\ \ \scriptsize(2 tasks) & 74.7 & 77.3 & 72.4 & 71.2 & 71.5 & 74.4 & 76.7 & 75.1 & 80.3 & \textbf{80.9} \\[1pt]
MSCOCO\_i2t & 71.1 & 74.6 & 69.1 & 69.8 & 68.9 & 75.3 & 71.4 & 76.3 & 76.9 & \textbf{77.7} \\
VisualNews\_i2t & 78.3 & 80.1 & 75.8 & 72.7 & 74.2 & 73.6 & 82.0 & 73.9 & 83.8 & \textbf{84.1} \\
\addlinespace[2pt]
\rowcolor{plmgray}\textbf{Image-to-image retrieval}\ \ \scriptsize(8 tasks) & 64.8 & \textbf{74.3} & 64.8 & 66.0 & 64.2 & 70.1 & 69.7 & 73.2 & 72.4 & 74.1 \\[1pt]
CIRR & 57.3 & \textbf{67.0} & 54.0 & 54.6 & 53.7 & 47.6 & 55.3 & 61.2 & 52.5 & 54.9 \\
FashionIQ & 19.6 & 27.0 & 19.1 & 20.3 & 17.1 & 24.0 & 23.4 & \textbf{31.9} & 21.2 & 26.7 \\
MSCOCO & 66.2 & 81.3 & 66.4 & 66.9 & 69.5 & 92.9 & 72.7 & \textbf{93.6} & 92.3 & 91.2 \\
NIGHTS & 68.4 & 68.3 & 68.1 & 68.0 & 67.2 & 66.3 & 68.1 & 66.4 & 67.9 & \textbf{69.2} \\
OVEN & 64.8 & 68.5 & 67.8 & 68.4 & 66.9 & 61.4 & \textbf{71.4} & 60.7 & 68.0 & 71.0 \\
RefCOCO & 87.0 & 95.3 & 87.2 & 86.5 & 83.3 & 94.9 & 91.4 & 95.9 & 96.1 & \textbf{96.6} \\
RefCOCO-Matching & 86.3 & \textbf{92.8} & 86.4 & 88.4 & 84.4 & 85.8 & 91.1 & 88.0 & 91.3 & 92.7 \\
Visual7W-Pointing & 69.0 & \textbf{94.0} & 69.4 & 74.9 & 71.5 & 88.0 & 84.2 & 87.9 & 89.8 & 90.4 \\
\addlinespace[2pt]
\rowcolor{plmgray}\textbf{Video classification}\ \ \scriptsize(5 tasks) & 39.2 & 37.2 & 43.7 & 45.0 & 44.3 & 57.0 & 48.6 & 57.6 & 53.8 & \textbf{58.1} \\[1pt]
Breakfast & 14.8 & 18.0 & 18.0 & 20.1 & 20.1 & \textbf{36.7} & 21.5 & 33.0 & 30.0 & 36.5 \\
HMDB51 & 40.2 & 42.8 & 47.1 & 51.2 & 54.4 & 56.7 & 58.3 & \textbf{60.1} & 47.5 & 54.4 \\
Kinetics-700 & 38.2 & 38.0 & 43.1 & 42.2 & 35.8 & 55.8 & 42.8 & \textbf{56.8} & 51.6 & 55.6 \\
SmthSmthV2 & 43.0 & 25.2 & 41.0 & 44.8 & 44.1 & 56.7 & 50.4 & 59.5 & 59.4 & \textbf{61.8} \\
UCF101 & 60.0 & 61.8 & 69.3 & 66.5 & 67.2 & 79.3 & 70.0 & 78.5 & 80.7 & \textbf{82.1} \\
\addlinespace[2pt]
\rowcolor{plmgray}\textbf{Video QA}\ \ \scriptsize(5 tasks) & 34.7 & 50.6 & 47.0 & 52.3 & 50.9 & 55.9 & 60.7 & 58.4 & 65.9 & \textbf{67.5} \\[1pt]
ActivityNetQA & 53.0 & 64.6 & 64.8 & 69.2 & 57.8 & 74.8 & 76.0 & 74.4 & 77.4 & \textbf{80.8} \\
EgoSchema & 35.0 & 51.6 & 37.0 & 47.8 & 45.4 & 53.0 & 52.4 & 52.8 & 60.4 & \textbf{60.6} \\
MVBench & 33.6 & 42.2 & 45.4 & 47.4 & 49.9 & 50.8 & 58.2 & 55.9 & 63.0 & \textbf{64.8} \\
NExTQA & 20.9 & 58.8 & 47.9 & 57.3 & 60.0 & 53.9 & 69.6 & 58.2 & 73.7 & \textbf{74.9} \\
Video-MME & 30.8 & 35.8 & 39.9 & 40.0 & 41.7 & 47.1 & 47.3 & 50.5 & 55.0 & \textbf{56.6} \\
\addlinespace[2pt]
\rowcolor{plmgray}\textbf{Video retrieval}\ \ \scriptsize(5 tasks) & 28.4 & 28.9 & 33.0 & 33.5 & 32.9 & 45.1 & 38.2 & 45.1 & 45.3 & \textbf{47.3} \\[1pt]
DiDeMo & 30.0 & 31.5 & 33.0 & 32.7 & 32.4 & 45.3 & 40.0 & 46.8 & 48.9 & \textbf{50.9} \\
MSR-VTT & 27.8 & 27.6 & 33.8 & 36.2 & 34.3 & 45.7 & 38.9 & 46.2 & 45.9 & \textbf{47.4} \\
MSVD & 47.3 & 57.5 & 56.0 & 56.1 & 55.4 & 67.2 & 60.7 & 65.8 & 66.9 & \textbf{67.5} \\
VATEX & 26.2 & 22.5 & 27.6 & 28.2 & 29.9 & \textbf{43.6} & 32.6 & 43.4 & 40.5 & 42.5 \\
YouCook2 & 10.6 & 5.6 & 14.5 & 14.5 & 12.7 & 23.5 & 18.5 & 23.3 & 24.6 & \textbf{28.1} \\
\addlinespace[2pt]
\rowcolor{plmgray}\textbf{Moment retrieval}\ \ \scriptsize(3 tasks) & 37.5 & 39.6 & 33.6 & 46.7 & 39.7 & 49.4 & 39.3 & \textbf{49.5} & 45.8 & 49.4 \\[1pt]
Charades-STA & 20.1 & 30.0 & 18.0 & 19.4 & 20.4 & 26.4 & 21.9 & 25.0 & 21.3 & \textbf{30.7} \\
MomentSeeker & 42.9 & 32.0 & 40.4 & \textbf{63.5} & 41.2 & 50.9 & 41.1 & 49.9 & 49.3 & 48.6 \\
QVHighlight & 49.7 & 56.8 & 42.2 & 57.1 & 57.5 & 70.7 & 54.8 & \textbf{73.6} & 66.8 & 69.1 \\
\addlinespace[2pt]
\rowcolor{plmgray}\textbf{Visual-document retrieval}\ \ \scriptsize(24 tasks) & 68.7 & 60.1 & 62.7 & 67.5 & 67.3 & 74.1 & 70.6 & 74.7 & 81.0 & \textbf{82.2} \\[1pt]
MMLongBench-doc & -- & 40.8 & 41.4 & 32.0 & 39.7 & 50.3 & 41.3 & 50.7 & 86.1 & \textbf{86.4} \\
MMLongBench-page-fixed & 44.7 & -- & -- & 39.9 & -- & 47.7 & -- & 51.0 & 71.4 & \textbf{72.3} \\
ViDoRe\_arxivqa & 78.9 & 51.9 & 76.5 & 72.6 & 73.9 & 86.1 & 73.6 & 88.7 & 90.9 & \textbf{91.2} \\
ViDoRe\_biomedical\_lectures\_v2\_multilingual & 44.6 & 33.8 & 39.1 & 48.2 & 46.1 & 51.0 & 50.7 & 50.1 & 59.7 & \textbf{61.3} \\
ViDoRe\_docvqa & 37.1 & 38.2 & 37.2 & 36.2 & 37.9 & 45.7 & 41.1 & 47.5 & 54.0 & \textbf{54.1} \\
ViDoRe\_economics\_reports\_v2\_multilingual & 42.3 & 36.0 & 39.7 & 49.6 & 45.7 & 53.0 & 57.8 & 53.9 & 61.9 & \textbf{65.4} \\
ViDoRe\_esg\_reports\_human\_labeled\_v2 & 45.8 & 54.7 & 40.0 & 52.1 & 50.2 & 56.9 & 50.4 & 59.8 & 58.0 & \textbf{68.7} \\
ViDoRe\_esg\_reports\_v2\_multilingual & 45.7 & 43.6 & 35.7 & 49.0 & 42.6 & 46.9 & 43.2 & 49.7 & 51.5 & \textbf{55.4} \\
ViDoRe\_infovqa & 82.7 & 73.2 & 80.9 & 79.0 & 76.2 & 86.8 & 80.8 & 86.9 & 91.5 & \textbf{91.5} \\
ViDoRe\_shiftproject & 61.0 & 45.4 & 62.3 & 64.8 & 66.8 & 70.7 & 65.0 & 69.0 & 78.7 & \textbf{79.0} \\
ViDoRe\_syntheticDocQA\_artificial\_intelligence & 89.1 & 76.8 & 79.9 & 83.8 & 85.9 & 94.0 & 89.5 & 91.6 & 96.3 & \textbf{96.5} \\
ViDoRe\_syntheticDocQA\_energy & 86.3 & 77.3 & 85.3 & 82.6 & 83.3 & 86.7 & 85.7 & 88.1 & 91.5 & \textbf{93.2} \\
ViDoRe\_syntheticDocQA\_government\_reports & 85.6 & 79.9 & 81.0 & 83.2 & 82.6 & 89.0 & 89.8 & 90.7 & 94.4 & \textbf{94.8} \\
ViDoRe\_syntheticDocQA\_healthcare\_industry & 91.1 & 81.7 & 85.3 & 91.1 & 90.8 & 91.1 & 94.3 & 90.4 & \textbf{95.5} & \textbf{95.6} \\
ViDoRe\_tabfquad & 87.8 & 57.7 & 80.4 & 88.8 & 86.1 & 94.5 & 90.2 & 94.7 & 94.4 & \textbf{95.5} \\
ViDoRe\_tatdqa & 44.3 & 35.5 & 42.0 & 36.6 & 40.6 & 54.6 & 46.7 & 54.8 & 62.8 & \textbf{63.9} \\
ViDoSeek-doc & -- & 75.8 & 78.3 & 76.9 & 75.9 & 82.6 & 75.3 & 82.4 & 89.0 & \textbf{89.1} \\
ViDoSeek-page-fixed & 80.3 & -- & -- & 80.6 & -- & 82.0 & -- & 84.4 & \textbf{84.5} & 83.8 \\
VisRAG\_ArxivQA & 76.7 & 53.1 & 76.9 & 71.6 & 74.3 & 84.9 & 80.5 & 86.9 & 88.9 & \textbf{89.7} \\
VisRAG\_ChartQA & 84.2 & 83.7 & 86.2 & 80.8 & 86.0 & 88.3 & 84.9 & 88.5 & 87.4 & \textbf{90.9} \\
VisRAG\_InfoVQA & 85.9 & 82.6 & 88.7 & 85.7 & 84.4 & 90.0 & 89.2 & 89.6 & 94.2 & \textbf{94.4} \\
VisRAG\_MP-DocVQA & 71.8 & 66.4 & 78.6 & 74.9 & 75.6 & 79.1 & 83.4 & 79.3 & \textbf{89.2} & 89.1 \\
VisRAG\_PlotQA & 65.9 & 51.0 & 65.6 & 66.2 & 68.0 & 73.0 & 72.7 & 72.4 & \textbf{75.3} & 74.2 \\
VisRAG\_SlideVQA & 91.4 & 86.4 & 91.8 & 88.9 & 87.1 & 92.3 & 91.5 & 92.6 & 95.8 & \textbf{96.3} \\
\end{longtable}
}

\end{document}